\documentclass[twoside,11pt]{article}
\usepackage{jair,rawfonts}
\usepackage{theapa}
\usepackage{amsmath}
\usepackage{url}
\usepackage{linguex}
\usepackage{multirow}
\usepackage{tikz}
\usepackage{qtree}
\let\chapter\section
\usepackage[]{algorithm2e}
\usepackage{float}
\usepackage{pgfplots}
\usepackage{rotating}
\usepackage{tikz}
\usepackage{tikz-qtree}
\newcommand{\amt}[1]{\textrm{\fontsize{8pt}{.1pt}\selectfont{#1}}}
\renewcommand\vec[1]{\overrightarrow{#1}}
\newcommand\cev[1]{\overleftarrow{#1}}
\ShortHeadings{
  Building a Neural Semantic Parser from a Domain Ontology} {Cheng, Reddy, \& Lapata}
\firstpageno{1}

\title{Building a Neural Semantic Parser from a Domain Ontology}

\author{\name Jianpeng Cheng \email jianpeng.cheng@ed.ac.uk \\
       \addr University of Edinburgh
       \AND
       \name Siva Reddy \email siva.reddy@cs.stanford.edu \\
       \addr Stanford University
       \AND
       \name Mirella Lapata \email mlap@inf.ed.ac.uk \\
       \addr University of Edinburgh
    }

\begin{document}

\maketitle

\begin{abstract}
  Semantic parsing is the task of converting natural language
  utterances into machine interpretable meaning representations which
  can be executed against a real-world environment such as a
  database. Scaling semantic parsing to arbitrary domains faces two
  interrelated challenges: obtaining broad coverage training data
  effectively and cheaply; and developing a model that generalizes to
   compositional utterances and complex intentions.  
  We
  address these challenges with a framework which allows to elicit
  training data from a domain ontology and bootstrap a neural parser which recursively builds derivations of logical forms.  In our framework meaning
  representations are described by sequences of natural language
  templates, where each template corresponds to a decomposed fragment
  of the underlying meaning representation.  Although artificial,
  templates can be understood and paraphrased by humans to create
  natural utterances, resulting in parallel triples of utterances,
  meaning representations, and their decompositions.  These allow us
  to train a neural semantic parser which learns to compose rules in
  deriving meaning representations.  We crowdsource training data on
  six domains, covering both single-turn utterances which exhibit
  rich compositionality, and sequential utterances where a complex
  task is procedurally performed in steps. We then develop neural semantic
  parsers which perform such compositional tasks.  In general, our approach allows to deploy neural
  semantic parsers quickly and cheaply from a given domain ontology.

\end{abstract}

\section{Introduction}

\begin {table}[t]
\begin{center}
	\footnotesize
	\begin{tabular}{|@{~}c@{~}|@{~}p{4cm}@{~}@{~}p{4cm}@{~}@{~}p{5.7cm}@{~}|}
     \hline
	{Task} & \multicolumn{1}{@{~}c@{~}}{Aggregated NL
          Utterances} & \multicolumn{1}{@{~}c@{~}}{Decomposed NL utterances} & \multicolumn{1}{c|}{{Meaning
            Representations}} \\ \hline \hline
	\raisebox{-2.8cm}[0pt]{\rotatebox{90}{Querying a database\hspace*{1cm}}} &
        \textit{\newline \newline Find all Chinese restaurants near me. My maximum budget is 50\$ and list only those with restrooms. }
	& \textit{Find all Chinese restaurants near
          me. \newline\newline Which ones cost no more than 50\$? \newline\newline And have restrooms?}
	& \texttt{r1 = filter$_{=}$(find\_all(restaurants), food\_type, chinese)  \newline  r2 = filter$_{<}$(r1, distance, 500m)
	\newline \newline r3 = filter$_{\leq}$(r2, price, 50\$)
        \newline \newline  r4 = filter\_assertion(r3, has\_restroom)
      }\\ \hline \hline

	\raisebox{-2.8cm}[0pt]{\rotatebox{90}{Instructing a robot}} &
        \textit{\newline\newline Place the kettle under the tap and fill it with water. When it is filled, turn off the tap and heat the kettle with the stove, until the water is boiled.} & 
	\textit{Place the kettle in the sink and fill it with water. \newline\newline Turn off the tap and heat the kettle with the stove. 
		\newline\newline Wait until the water is boiled. }
	& \texttt{move(kettle, sink)  \newline toggle(sinkknob, on)  \newline \newline wait\_until(fill) \newline toggle(sinknob, off) \newline   move(kettle, stove) 
		\newline toggle(stoveknob, on) \newline\newline wait\_until(boil) \newline toggle(stoveknob, off) } \\ \hline
	\end{tabular}
\end{center}
\caption{Examples of natural language (NL) utterances and corresponding
  meaning representations.  Human intentions may be specified as
  a longer and more compositional utterance, or a sequence of inter-related short utterances.\label{examples}} 
\end{table}


Semantic parsing has recently emerged as a key technology towards
developing systems that understand natural language and enable
interactions between humans and computers.  A semantic parser converts
natural language utterances into machine interpretable meaning
representations, such as programs or logical forms. These representations can be executed against a real-world environment, such as
navigating a database, comparing products, or controlling a robot
\cite{kwiatkowski2011lexical,liang2011learning,wen2015semantically,matuszek2013learning}.  Table~\ref{examples} shows
examples of natural language utterances and their corresponding
meaning representations in two application scenarios.

The development of semantic parsers faces two interrelated challenges,
namely how to obtain training data efficiently and cheaply and how to
handle compositional utterances and intentions.  
As far as the first
challenge is concerned, semantic parsers have been mostly trained on
data consisting of utterances paired with human-annotated meaning
representations
\cite{zelle1996learning,zettlemoyer_learning_2005,wong:learning:2006,kwiatkowksi2010inducing,dong2016language,jia2016data}. Labeling
such data is labor-intensive and error-prone: annotators must not only be trained with the knowledge about the logical language and the domain of interest, but also need to ensure every meaning representation they create actually matches the utterance
semantics.   
Because of this reason, it tasks much effort to develop neural semantic parsers for new domains where no training data exists; and for domains
whose ontology changes frequently (training data needs to be updated accordingly).
This annotation challenge becomes more obvious when human intentions become complex.
Table~\ref{examples} shows examples of complex
intentions expressed as a single compositional utterance
or as a sequence of simpler utterances.   In either case, writing down the correct meaning representations is not trivial.

As for the second challenge, traditional semantic parsers
\cite{zettlemoyer:learning:2005,wong:learning:2006,kwiatkowksi2010inducing,kwiatkowski2013scaling,berant2013semantic}
adopt a domain-specific grammar, a trainable model, and a parsing
algorithm. The grammar defines the space of possible derivations from an utterance to a logical form, and the model together
with the parsing algorithm finds the most likely derivation.  A chart-based
parsing algorithm is commonly used to parse an utterance in polynomial
time. Recent advances in neural networks have spurred new interest in
reformulating semantic parsing as a sequence-to-sequence learning
problem \cite{bahdanau2014neural}.  Such neural semantic parsers
\cite{dong2016language,jia2016data} parse utterances in linear time,
while reducing the need for grammar and feature engineering. But this modeling
flexibility comes at a cost since it is less possible to
interpret how meaning composition is performed---because meaning
representations are treated as strings rather than structured objects
(e.g.,~trees or graphs). Such knowledge plays a critical role in
building more generalizable neural semantic parsers, especially in the face
of sparse data.  For this reason, subsequent development of neural semantic parsers focuses on sequence-to-action models which handle meaning composition explicitly \cite{cheng2017learning,cheng2017nsp,yin2018tranx,gupta2018semantic}. 
Besides, most previous work on
neural semantic parsing has focused on isolated utterances, ignoring
the fact that some intentions are more likely to be expressed
 through a sequence of co-referring utterances (see
Table~\ref{examples}).

Our first contribution in this paper is a method for eliciting neural
semantic parsing data which expresses complex intentions, from a domain ontology.  Our approach builds on the work of
\shortciteA{wang2015building} who advocate crowd-sourcing as a way
of mitigating the paucity of semantic parsing datasets. Their basic
idea is to use a synchronous grammar to generate meaning
representations paired with artificial utterances which crowdworkers
are asked to paraphrase into more natural sounding utterances.  For example,
\texttt{argmin(food\_type(Thai food), distance)} is deterministically
mapped to ``\textsl{restaurants with smallest distance where food type
  is thai}", which will be later paraphrased by crowdworkers into
e.g.,~``\textsl{nearest restaurants serving thai food}''.  However, we experimentally found out
the readability of these artificial utterances decreases when the
complexity of the task increases.  As an example, the artificial
utterance ``\textsl{restaurants where food type is food type of kfc
  which has minimum price}''  is rather difficult to interpret
due to the attachment ambiguity caused by relative clause
\textsl{``which has minimum price''}---it is unclear whether it
modifies \textsl{kfc}, \textsl{food}, or \textsl{restaurants}. The
approach primarily targets utterances exhibiting shallow
compositionality often with two predicates and entities
\cite{wang2015building}, thereby avoiding ambiguities arising from
complex intentions.
Instead of representing the meaning of a  task with a single
artificial description, our approach represents it as a
sequence of inter-related templates, where each template corresponds
to a decomposed fragment of meaning.  So, the example above would be represented
with templates ``\texttt{Result$_1$ = find food type kfc}'' and
``\texttt{Result$_2$ = find restaurants with food type Result$_1$}''
and ``\texttt{Result$_3$ = find Result$_2$ with minimum price}''.
\shortciteA{iyyer2017search} show that decomposed utterances can
capture higher levels of compositionality in practice.  Since templates correspond
to specific parts of the meaning representation, they are easier to
understand by crowdworkers compared to more elaborate artificial
descriptions. Furthermore, the templates allow us to flexibly
crowdsource two different types of data
to study human intentions expressed in  different ways:
we can obtain individual utterances which
correspond to more compositional meaning representations,  or  a
sequence of inter-related utterances, depending on whether
participants are asked to summarize or  paraphrase the templates.

The second contribution of this paper is a neural semantic parsing
framework which leverages the annotations (in the form of template sequences) elicited by the above method
and the ability of recurrent neural
networks to model compositionality
\cite{dyer2016recurrent,cai2017making}. Our model is based on the fact that a
sequence of templates encodes the rules whose
application in recursive order yields the final meaning
representation. The parser is thus trained to predict derivations\footnote{In this work, a derivation tree refers to a parse tree that graphically represents the semantic information of how a meaning representation is derived from a context-free grammar, which does not reply on tokens in the corresponding utterance---this is slightly different from the definition of derivation in a chart parser.},
obtaining meaning representations by composing the rules. Specifically,
we adopt a transition action-based approach which handles the generation of
domain-general and domain-specific rules in a unified way, with
constraints ensuring the rules can be composed smoothly.  An important
challenge in semantic parsing is tackling mismatches between natural
language and representation language.  For example, both utterances
``\textsl{cheapest restaurants}'' and ``\textsl{restaurants with
  smallest price rating}'' trigger an \texttt{argmax} rule despite
expressing the same information need in different ways. We resolve
this challenge with a neural attention mechanism, which learns a soft
mapping between natural language and meaning representation language.
The neural semantic parser is also designed to handle sequential
utterances\footnote{This work studies parsing sequential utterances in a non-dialog setup: the model does not involve decision making on the optimum strategy of responding to each input utterance. Instead, it simply outputs the execution result of the obtained meaning representation.} which involve co-reference.

We conduct a wide range of experiments to evaluate the proposed
framework.  As a testbed, we elicit annotations for database querying tasks
involving compositional user intentions.  Using crowdsourcing, various templates underlying
computer-generated meaning representations are labeled with either
single or sequential utterances.  We crowdsource data
covering six domains and provide detailed analysis on the annotations. We then use the data to train a
neural semantic parser which handles compositionality and
co-reference.  Overall, we advocate an end-to-end solution which
allows to build neural semantic parsers quickly and cheaply, starting with a domain ontology.

The remainder of this paper is structured as
follows. Section~\ref{rw} discusses related work of semantic parsing.
Section~\ref{data} introduces our data elicitation method while
Section~\ref{nsp} presents the neural semantic parsing model. 
Section~\ref{exp} highlights our experimental results.  Finally, Section~\ref{conclusion}
concludes the paper.

\section{Related Work \label{rw}}
\label{sec:related-work}
Early semantic parsing systems are hard-coded to answer questions in constrained domains.
The \textsc{lunar} system \cite{winograd1972understanding} were designed to handle questions about moon rocks using a large database. 
It converts queries into programs by mapping syntactic fragments to semantic units.
Another example is the \textsc{shrdlu} system   \cite{winograd1972understanding} which launches dialogs between the user and the system-simulated robot to manipulate simple objects on a table. 
Central to these systems is the idea of expressing words and sentences as computer programs,
and the execution of programs corresponds to the reasoning of meanings.
However, the development of these systems
require a large deal of  domain-specific knowledge and engineering.

In reaction to these problems in 1970s, the focus of semantic parsing research shifted from rule-based method to
empirical or statistical methods, where data and machine learning plays an important role. 
Statistical
semantic parsers typically consist of three key components: a grammar, a
trainable model, and a parsing algorithm. The grammar defines the
space of derivations from utterances to meaning representations, and the model
together with the parsing algorithm find the most likely
derivation. 
An example of early statistical semantic parser is the \textsc{chill} system \cite{zelle1996learning}
based on inductive logic programming (ILP).
The system uses ILP to learn control rules for a shift-reduce parser.
To train and evaluate their system, \shortciteA{zelle1996learning} created the \textsc{geoquery} dataset which contains 880 queries to a US geography database.
These queries are paired with annotated meaning representations in Prolog.

Until early 2000, semantic parsing research mainly focused on restricted domains.
Besides \textsc{geoquery},  commonly used datasets are \textsc{robocup} for coaching advice to soccer agents \cite{kitano1997robocup}, and \textsc{atis} for air travel information service \cite{price1990evaluation}.
At that time, statistical approaches for parsing domains-specific context-free grammars have been largely explored.
For example, \shortciteA{kate2006using} propose \textsc{krisp}, which induces context-free grammar rules that generate meaning representations, and uses kernel SVM to score derivations.
\shortciteA{ge2005statistical} propose \textsc{scissor},
which employs
an integrated statistical parser to produce
a semantically augmented parse tree.
Each non-terminal node in the tree has
both a syntactic and a semantic label, from which the final meaning representation can be derived.
The \textsc{wasp} system proposed by \shortciteA{wong2007learning} 
learns synchronous context free grammars that generate utterances and meaning representations. Parsing is achieved by 
finding the most probable derivation that leads to the utterance and recovering the meaning representation with synchronous rules. 
\shortciteA{lu2008generative} proposes a generative model for utterances and meaning representations.
Similar to \shortciteA{ge2005statistical}, they define hybrid trees whose nodes include both words and meaning representation tokens. 
Training is performed with the EM algorithm. The model, especially the generative process, was extend by \shortciteA{kim2010generative} to learn from ambiguous supervisions. 

The next breakthrough came with the work of \shortciteA{zettlemoyer:learning:2005}, who introduced CCG in semantic parsing.
Their probabilistic CCG grammars can deal with long range dependencies and construct non-projective meaning representations.
A great deal of work follows \shortciteA{zettlemoyer:learning:2005} but focuses on more fine-grained problems such as grammar induction and lexicon learning \cite{kwiatkowski2010inducing,kwiatkowski2011lexical,krishnamurthy2012weakly,artzi2014learning,krishnamurthy2015learning,krishnamurthy2016probabilistic,gardner2017open}
or using less supervision \cite{artzi-zettlemoyer:2013:TACL,reddy2014large}.
As a common paradigm, the class of work first generates candidate derivations to meaning representations governed by the grammar. 
These candidates derivations are scored by a trainable model which can take the form of a structured
perceptron \cite{zettlemoyer2007online} or a log-linear model
\cite{zettlemoyer:learning:2005}. 
Training updates model parameters such that good derivations obtain higher scores. 
During inference, a CKY-style
chart parsing algorithm is used to predict the most likely derivation
for an utterance. 
Another class of work follows similar paradigm but use lambda DCS as the semantic formalism \cite{berant-EtAl:2013:EMNLP,berant2014semantic,berant2015imitation}. 
Other interesting work includes joint semantic parsing and grounding \cite{kwiatkowski2013scaling}, 
parsing context-dependent queries \cite{artzi2013weakly,long2016simpler},
and converting dependency trees to meaning representations \cite{reddy2016transforming,reddy2017universal}.

With recent advances in neural networks and deep learning, 
there is a trend of reformulating semantic parsing as a machine translation problem,
which converts a natural language sequence into a programming language consequence. 
The idea was not novel and has been previously studied with statistical machine translation
approaches. For example, both \shortciteA{wong:learning:2006} and \shortciteA{andreas_semantic_2013}  developed word-alignment based translation models for parsing the \textsc{geoquery} dataset.
However, the task setup is important to be revisited since recurrent neural networks 
have been shown to be extremely useful in context modeling and sequence generation \cite{bahdanau2014neural}. 
Following this direction, \shortciteA{dong2016language} and \shortciteA{jia2016data} developed neural semantic parsers which treat semantic parsing as 
a sequence-to-sequence learning problem. Surprisingly, the approach has been proven effectively on even the small \textsc{geoquery} dataset.
\shortciteA{jia2016data}  further introduces a data augmentation approach which
 bootstraps a synchronous grammar from existing data and generates artificial examples as extra training data. 
State of the art result on \textsc{geoquery} dataset was obtained with this approach.
Subsequent work of \shortciteA{kovcisky2016semantic} attempts to explore the meaning representation space with a generative autoencoder. 
They bootstraps a probabilistic monolingual grammar for meaning representations, from which unseen meaning representations can be sampled.
These samples are used as semi-supervised training data to the autoencoder. 
Other related work extends the vanilla sequence to sequence model in various ways, 
such as employing two encoder-decoders for coarse to fine decoding \cite{dong2018coarse},
handling multiple tasks with a shared encoder \cite{fan2017transfer},
parsing cross-domain queries \cite{herzig2017neural} and context-dependent queries \cite{suhr2018learning}, 
and applying the model to other formalisms such as AMR \cite{konstas2017neural} and SQL  \cite{zhong2017seq2sql,xu2017sqlnet}.

The fact that meaning representations have a syntactic structure has motivated more recent work
on exploring structured neural decoders to generate tree or graph structures, 
and grammar constrained decoders to make sure the outputs are meaningful and executable. 
For example, \shortciteA{yin2017syntactic}  generate abstract syntax trees for source code with a grammar constrained neural decoder. 
\shortciteA{krishnamurthy2017neural} also introduce a neural semantic parser which decodes rules of a grammar to obtain well-typed meaning representations.
\shortciteA{cheng2017learning,cheng2017nsp,yin2018tranx,gupta2018semantic,chen2018sequence} all employ neural sequence-to-action models to generate structured meaning representations.

\section{Data Elicitation \label{data}}
As shown above, most work on semantic parsing (including neural semantic parsing) has used existing datasets with annotated
 utterance-meaning
representation pairs 
\cite{zelle1996learning,ge2005statistical,kate2006using,kate2005learning,wong:learning:2006,lu2008generative,kwiatkowksi2010inducing,dong2016language,jia2016data}.
In contrast, our work focuses on a practical scenario when one wants to develop a neural semantic parser from a new domain ontology: there exists no prior training data but the expected utterances can be arbitrarily compositional.
Two challenges arise here for data collection: 1) although training data in form of  utterance-meaning
representation pairs provides an effective training signal,  their annotation is labor intensive and error-prone.
2) although utterances (e.g., usage logs if exist) can be sampled for the given domain, it is not easy to ensure this data covers a broad range of compositional patterns.

In this section, we
detail a data elicitation method which collects training data of neural semantic
parsers cheaply and effectively, for both single-turn and
sequential utterances. The idea is to use a computer program to generate valid meaning representations based on the domain ontology.
These meaning representations are mapped to an artificial human language, which can be understood by annotators to create utterances. 
In summary, annotators are generating utterances for meaning representations, instead of generating meaning representations for utterances.
 
\subsection{Decomposition of Meaning Representations \label{decom}}

Our approach follows \shortciteA{wang2015building} to
decompose meaning representations into various constructs.  The
decomposition allows us to build a program which generates meaning
representations with broad coverage; and explicitly model the
generation process during parsing.  Throughout this paper, we
exemplify our approach with a database querying task.
Specifically, our meaning representations are written in lambda expressions
representing rules and variables in a computer program that queries a
database.

\begin{table}[!ht]
	\begin{center}
		\small
		\begin{tabular}{| @{~}p{3cm}@{~} @{~}p{5cm}@{~}@{~}p{6cm}|}
                  \hline 
                  \multicolumn{1}{|c@{~}@{~}}{Category}  & \multicolumn{1}{c@{~}@{~}}{Domain-general Rules} & \multicolumn{1}{c|}{Description and Evaluation}  \\ \hline\hline
                  \texttt{LookupKey} & \texttt{$\lambda$s:(lookupKey (var s))} & Looks for the entire set of $s$ \\ \hline
                  \texttt{LookupValue} & \texttt{$\lambda$p$\lambda$s:(lookupValue (var s) (var p))} & Looks for  specific property $p$ of  entity $s$ \\ \hline
                  \texttt{Filter(property)} &  \texttt{$\lambda$s$\lambda$p$\lambda$v:(filter (var s) (var p) = (var v))} & Looks for subset of $s$ whose property $p$ equates to some value~$v$ \\ \hline
                  \texttt{Filter(assertion)}& 
                  \texttt{$\lambda$s$\lambda$p:(filter (var s) (var p) = true)} &  Looks for subset of $s$ which satisfies condition $p$ \\ \hline
                  \texttt{Count}  &  \texttt{$\lambda$s:(size (var s))} & Computes  total number of elements in  set $s$  \\ \hline
                  \texttt{Sum} &   \texttt{$\lambda$s:(sum (var s))} & Computes total sum of numerical elements in  set $s$   \\ \hline
                  \texttt{Comparative} ($<$) &  \texttt{$\lambda$s$\lambda$p$\lambda$v:(filter (var s) (var p) $<$ (var v))} & Looks for subset of $s$ whose numeric property $p$ is smaller than some numeric value $v$\\ \hline
                  \texttt{Comparative} ($\leq$) &  \texttt{$\lambda$s$\lambda$p$\lambda$v:(filter (var s) (var p) $\leq$ (var v))} & Looks for subset of $s$ whose numeric property $p$ is smaller than or equal to some numeric value $v$\\ \hline
                  \texttt{Comparative} ($>$) &  \texttt{$\lambda$s$\lambda$p$\lambda$v:(filter (var s) (var p) $>$ (var v))} & Looks for subset of $s$ whose numeric property $p$ is larger than some numeric value $v$\\ \hline
                  \texttt{Comparative} ($\geq$) &  \texttt{$\lambda$s$\lambda$p$\lambda$v:(filter (var s) (var p) $\geq$ (var v))} & Looks for subset of $s$ whose numeric property $p$ is larger than or equal to some numeric value $v$\\ \hline
                  \texttt{CountComparative} ($<$) & 
                  \texttt{$\lambda$s$\lambda$p$\lambda$v:((var s) (size (var p)) $<$ (var v))} &  Looks for subset of $s$ where the cardinality of property $p$ is smaller than some numeric value $v$ \\ \hline
                  \texttt{CountComparative} ($\leq$) & 
                  \texttt{$\lambda$s$\lambda$p$\lambda$v:((var s) (size (var p))  $\leq$ (var v))} &  Looks for subset of $s$ where the cardinality of property $p$ is smaller than or equal to some numeric value $v$ \\ \hline
                  \texttt{CountComparative} ($>$) & 
                  \texttt{$\lambda$s$\lambda$p$\lambda$v:((var s) (size (var p))  $>$ (var v))} &  Looks for subset of $s$ where the cardinality of property $p$ is larger than some numeric value $v$ \\ \hline
                  \texttt{CountComparative} ($\geq$) & 
                  \texttt{$\lambda$s$\lambda$p$\lambda$v:((var s) (size (var p)) $\geq$ (var v))} &  Looks for subset of $s$ where the cardinality of property $p$ is larger than or equal to some numeric value $v$ \\ \hline
                  \texttt{Superlative} (min) & 
                  \texttt{$\lambda$s$\lambda$p:((var s) argmin (var p))} & Looks for  subset of $s$ whose numeric property $p$ is  smallest \\ \hline
                  \texttt{Superlative} (max) & 
                  \texttt{$\lambda$s$\lambda$p:((var s) argmax (var p))} & Looks for  subset of $s$ whose numeric property $p$ is largest \\ \hline
                  \texttt{CountSuperlative} (min) & 
                  \texttt{$\lambda$s$\lambda$p:((var s) argmin (size (var p)))} &  Looks  subset of $s$ where the cardinality of  property $p$ is smallest \\ \hline
                  \texttt{CountSuperlative} (max) & 
                  \texttt{$\lambda$s$\lambda$p:((var s) argmax (size (var p)))} &  Looks for  subset of $s$ where the cardinality of  property $p$ is largest \\
                  \hline
		\end{tabular}
	\end{center}
	\vspace{-2ex}
	\caption{Domain-general rules (and their descriptions) used to
          define meaning representations in our experiments. 
          \label{general-aspects}}
\end{table}

The first construct of meaning representations are
\emph{domain-general} rules stemming from the formal language used by
the semantic parser.  In Table~\ref{general-aspects} we provide
examples of domain-general rules represented as lambda
expressions. These rules specifying various functionalities such as
looking up a column in the database, counting, aggregation, and
filtering by condition.  They are generic, apply across domains, and
relevant to the database querying task.  The second construct of
meaning representations are \emph{domain-specific} rules which
generate domain-specific predicates or
entities. Table~\ref{specific-rule} shows example predicates and
entities (which are represented as variables in the formal language)
from the restaurant domain.  We are assuming access to a
domain-specific ontology which covers binary predicates for properties
(e.g.,~\texttt{custom\_rating}), unary predicates for assertions
(e.g.,~\texttt{open\_now}), and entities
(e.g.,~\texttt{restaurant.kfc}).

\begin {table}[t]
\begin{center}
	\small
	\begin{tabular}{|@{~}l@{}|@{~}l@{~}|@{~}l@{~}|}
          \hline
          \multicolumn{1}{|@{}c@{~}|@{~}}{Category} &
          \multicolumn{1}{c@{~}|@{~}}{Predicates and Entities} &
          \multicolumn{1}{c|}{Description}  \\ \hline \hline
          \multirow{7}{*}{\texttt{BinaryPredicate} } & \texttt{custom\_rating} & Overall rating from customers \\
          & \texttt{price\_rating} & Price rating from customers\\ 
          & \texttt{distance} & Distance of the restaurant\\ 
          & \texttt{num\_reviews} & Number of reviews from customers\\
          & \texttt{location} & Location of the restaurant\\ 
          & \texttt{cuisine} & Type of food served by the restaurant\\ 
          & \texttt{open\_time} & Opening time of the restaurant\\ \hline\hline
          \multirow{ 8}{*}{\texttt{UnaryPredicate}} &  \texttt{open\_now}  & Is the restaurant opening now? \\
          {} & \texttt{take\_away}  & Does the restaurant offer take-away?   \\
          {} & \texttt{reservation}  & Does the restaurant accept reservations?   \\
          {} & \texttt{credit\_card}  & Does the restaurant accept credit cards?   \\
          {} & \texttt{waiter}  & Does the restaurant have waiter service?   \\
          {} & \texttt{delivery}  & Does the restaurant offer delivery?  \\ 
          {} & \texttt{kids}  & Is the restaurant suitable for kids?  \\ 
          {} & \texttt{groups}  & Is the restaurant suitable for groups?   \\ \hline\hline
          \multirow{ 1}{*}{\texttt{Entity}} & \texttt{restaurant.kfc} & KFC \\
          {} & \texttt{location.oxford\_street}  & Oxford Street  \\ 
          \hline
	\end{tabular}
\end{center}
\caption{Domain-specific predicates and entities from a restaurant domain, covering binary
  predicates (properties), unary predicates (assertions), and
  entities.}  
\label{specific-rule}
\end{table}

The third construct of meaning representations captures complex human
intentions with co-referential variables, which establish anaphoric
links between meaning representations (antecedents and consequents).
To model co-reference, we adopt the notions of discourse referents
(DRs) and discourse entities (DEs) which are widepresed in discourse
representation theories \cite{webber1978formal,kamp2013discourse}. DRs
are referential expressions appearing in utterances which denote DEs,
i.e.,~mental entities in the speaker’s model of discourse.
Co-referential variables imply that DEs in the antecedent and
consequent refer to the same real-world entity (or entities) which we
obtain from the execution of the antecedent. In corresponding natural
language utterances, co-reference manifests itself by the explicit or
implicit occurrence of a pronoun or a DR (e.g.,~a definite noun
phrase) in the consequent.  The main principle in determining whether
DRs co-refer is that it must be possible to infer their relation from
the dialogue context alone, without using world knowledge.
Example~\ref{ex:1} below shows various continuations of the utterance
\textsl{Which restaurants serve thai food?} involving explicit
co-reference (i.e.,~\textsl{of those} in \ref{ex:2}), implicit
co-reference (see \ref{ex:3}), and co-reference via a definite
expression (i.e.,~\textsl{thai restaurants} in
\ref{ex:4}). All these co-references are meant to be represented by co-referential variables on the meaning representation side.
In this work, we consider three types of
co-referential variables shown in Table~\ref{coreft} which can be
used in place of type-matched, domain-specific entities to
construct meaning representations.

\ex. \label{ex:1} Which restaurants serve thai food?
   \a. \label{ex:2} Of those which ones are nearest to me? 
   \b. \label{ex:3} Nearest to me?
   \c. \label{ex:4} I mean thai restaurants nearest to me. 


\begin {table}[t]
\begin{center}
	\begin{tabular}{|l| l |}
		\hline
		\multicolumn{1}{|c}{Category} & \multicolumn{1}{|c|}{Description}  \\ \hline\hline
		\texttt{Coref} &  Refers to a single antecedent \\  
		\texttt{Union\_coref} & Refers to  the union of two antecedents\\  
		\texttt{Intersection\_coref} & Refers to the  intersection of two antecedents \\
		\hline
	\end{tabular}
\end{center}
\caption{Co-reference variables and their descriptions.}  
\label{coreft}
\end{table}

\subsection{Mapping Rules to Human Language}
\label{sec:mapp-rules-natur}

The central idea behind our data elicitation method is to convert
meaning representations to artificial descriptions,
by mapping the various rules used to construct meaning representations
to human language. The mapping enables the data elicitation procedures which will be described in Section \ref{sec:data collectiom}.

\begin{table}[t]
\begin{center}
	\small
	\begin{tabular}{|@{~}l@{~}| @{~}p{6cm}|}
		\hline
		\multicolumn{1}{|@{~}c@{~}|@{~}}{Category} &
                \multicolumn{1}{c|}{NL Templates}  \\
                \hline \hline
		\texttt{LookupKey}  & \textit{find all of \$s}  \\ 
		\texttt{LookupValue}  & \textit{find \$p of  \$s} \\ 
		\texttt{Filter(property)} & \textit{find \$s where \$p is \$v} \\ 
		\texttt{Filter(assertion)} & \textit{find \$s which satisfies \$p} \\ 
		\texttt{Count}  & \textit{count number of elements in \$s} \\ 
		\texttt{Sum}  & \textit{sum all elements in \$s} \\ 
		\texttt{Comparative($<$) }&  \textit{find \$s with \$p $<$ \$v} \\ 
		\texttt{Comparative($>$) }&  \textit{find \$s with \$p $>$ \$v} \\ 
		\texttt{Comparative($\leq$) }&  \textit{find \$s with \$p $\leq$ \$v} \\ 
		\texttt{Comparative($\geq$) }&  \textit{find \$s with \$p $\geq$ \$v} \\ 
		\texttt{CountComparative($<$) } & \textit{find \$s with  number of \$p $<$ \$v} \\ 
		\texttt{CountComparative($>$) } & \textit{find \$s with  number of \$p $>$ \$v} \\ 
		\texttt{CountComparative($\leq$) } & \textit{find \$s with  number of \$p $\leq$ \$v} \\ 
		\texttt{CountComparative($\geq$) } & \textit{find \$s with  number of \$p $\geq$ \$v} \\ 
		\texttt{Superlative(min)} & \textit{find \$s with smallest \$p } \\
		\texttt{Superlative(max)} & \textit{find \$s with largest \$p } \\
		\texttt{CountSuperlative(min)} & \textit{find \$s with smallest number of \$p} \\
		\texttt{CountSuperlative(max)} & \textit{find \$s with largest number of \$p} \\
		\hline
	\end{tabular}
\end{center}
\caption{Domain-general rules are associated with natural
	language (NL) templates specified by our
	framework.
	\label{general-template}}
\end{table}

Our method maps domain-general rules onto natural language
templates with missing entries.  Each template describes the
functionality of a rule, while missing entries specify variables
required by the rule.  Table~\ref{general-template} displays the
list of domain-general rules we use to query a database.  The
corresponding templates are described in such a way that can be
understood by annotators who have no knowledge of the underlying
meaning representation.  Different from the more natural language
descriptions shown in Table~\ref{general-aspects}, templates are human
readable formal descriptions which are deterministically mapped from meaning representations (see the right column in the table).

Domain-specific variables are mapped to natural language phrases with
a lexicon specified by a domain manager.  This lexicon is the only
resource we ask domain managers to provide, for the purposes of
describing the domain ontology.  Note that natural language
descriptions are important in cases where domain-specific predicates
or entities are not verbalized (for example a predicate may be simply
represented as an index \texttt{m.001} in the database). Such
descriptions must be provided to annotators to allow for basic
understanding, and to enable the paraphrasing task.  However, we do
not use this lexicon for building a semantic parser.
Table~\ref{lexicon} displays a lexicon for the restaurant domain.  
Natural language descriptions of predicates and entities are used to
instantiate templates.

\begin {table}[!ht]
\begin{center}
	\small
	\begin{tabular}{|l@{}|@{~}l|}
          \hline
          \multicolumn{1}{|c@{~}|@{~}}{Database Predicates/Entities} &
          \multicolumn{1}{c|}{NL Expressions}  \\ \hline \hline
          \texttt{custom\_rating} & customer rating \\
          \texttt{price\_rating} &  price rating \\ 
          \texttt{distance} & distance\\ 
          \texttt{num\_reviews} & number of customer reviews\\
	  \texttt{location} & location \\ 
          \texttt{cuisine} & cuisine \\ 
          \texttt{open\_time} & opening time \\
          \texttt{open\_now}  &  opens now \\
          \texttt{take\_away}  &  offers take-away   \\
          \texttt{reservation}  & takes reservations   \\
          \texttt{credit\_card}  & accepts credit cards \\
          \texttt{waiter}  & has waiter service   \\
          \texttt{delivery}  & offers delivery  \\ 
          \texttt{kids}  & suitable for kids  \\ 
          \texttt{groups}  & suitable for groups   \\ 
          \texttt{restaurant.kfc} & KFC \\
          \texttt{location.oxford\_street}  & Oxford Street  \\ 
          \hline
	\end{tabular}
\end{center}
\caption{Examples of domain-specific lexicon and corresponding natural
  language (NL) expressions for the restaurant domain.}  
\label{lexicon}
\end{table}

Finally, for co-referential variables, we directly assign an index
(e.g.,~\texttt{Result$_{1}$}, \texttt{Result$_{2}$}) to each fragment
of meaning representation, and use this index as the value of
co-referential variables when they are used to instantiate templates.

\subsection{Data Elicitation for Single-turn Utterances \label{sec:data collectiom}}
We are  now ready to describe our data elicitation procedures.  As mentioned earlier, our goal is to collect
utterance-meaning representation pairs which represent compositional intentions.
One intention can be
expressed within a single utterance, or a sequence of utterances. We
first discuss the data elicitation procedures for single-turn
utterances and then explain how it can be straightforwardly extended
to the sequential scenario.

 \begin {table}[!ht]
 \begin{center}
 	\small
 	\begin{tabular}{|lp{14cm}|}
          \hline
          1. & Bottom-up construction of meaning representations (done by framework) \\
          \multicolumn{2}{|c|}{} \\
          & 	   Domain-general rule
          \texttt{$\lambda$s:(lookupKey (var s))} is instantiated
          with domain-specific variable \texttt{type.restaurant}  
          to obtain the first piece of meaning representation: 

 	       \begin{center}
\texttt{Result$_{1}$=(lookupKey (type.restaurant))} 
\end{center} 

 	   For the second piece of meaning representation. 
domain-general rule
\texttt{$\lambda$s$\lambda$p$\lambda$v:(filter (var s)
  (var p) $=$ (var v))} is instantiated  with domain-specific and co-referential variables
\texttt{Result$_{1}$}, \texttt{rel.cuisine}, and 
\texttt{cuisine.thai}: 

 	       \begin{center}\texttt{Result$_{2}$=(filter (Result$_{1}$) (rel.cuisine) = (cuisine.thai))}\end{center} 
 	       
               For the third piece of meaning representation,
               domain-general rule
               \texttt{$\lambda$s:(lookupValue (var s) (var p))} is
               instantiated with
               domain-specific variables
               \texttt{restaurant.kfc} and \texttt{rel.distance}:
\begin{center}
 	      \texttt{Result$_{3}$=(lookupValue (restaurant.kfc) (rel.distance))} 
 \end{center}
 The final piece of meaning representation is constructed with
 domain-general rule \texttt{$\lambda$s$\lambda$p$\lambda$v:(filter (var s) (var p) $<$ (var v))}
 and domain-specific variables \texttt{Result$_{2}$},
 \texttt{rel.distance}, and \texttt{Result$_{3}$}:
\begin{center}
\texttt{Result$_{4}$=(filter (Result$_{2}$) (rel.distance) $<$ (Result$_{3}$))}\\
\end{center} \\\hline
2. &  Each piece of meaning representation is converted to a canonical
representation described by templates.  Templates associated with
domain-general rules are instantiated with
domain-specific and co-referential variables (shown in brackets): \\
\multicolumn{2}{|c|}{}\\

& \hspace{3cm}\textit{Result$_{1}$ = find all}  [\textit{restaurants}]\\

& \hspace{3cm}\textit{Result$_{2}$ = find} [\textit{Result$_{1}$}] \textit{where}
[\textit{cuisine}] \textit{is} [\textit{Thai}]  \\

& \hspace{3cm}\textit{Result$_{3}$ = find}  [\textit{distance}] of [\textit{KFC}]\\

& \hspace{3cm}\textit{Result$_{4}$ = find} [\textit{Result$_{2}$}] \textit{with} [\textit{distance}] \textit{$<$} [Result$_{3}$]  \\\hline 		
3. & The templates are displayed to the annotators who summarize them
into an utterance:
\begin{center}
\textit{Which restaurant has Thai food and is closer to me than KFC?} 
\end{center} \\

 		\hline
 	\end{tabular}
 \end{center}
 \caption{Example of our data elicitation procedures for single-turn
   utterances paired with meaning representations.}  
 \label{tab:single}
\end{table}

 \paragraph*{Step 1: Generating Meaning Representations} 
 We use a context-free grammar and  a computer program to generate meaning representations by
 sampling domain-general and specific rules.  Generation is performed
 in a bottom-up manner, so that larger pieces of meaning can be
 constructed from smaller ones.
 The application of each
 domain-general rule takes an expected amount (i.e. as defined by the grammar) of domain-specific or co-referential variables, and
 results in a new piece of meaning. Such bottom-up generation ensures the quality of meaning representations: the program can check whether the 
 generation process should continue with denotational clues.  Table \ref{tab:single} shows an example of the generation process.
 
Note that we do not consider complex
 co-referential structures between meaning representations in the
 single-turn case \footnote{In most of the cases, the next meaning representation co-refers to the previous one: \textsl{Result$_2$ = find Result$_1$ with
   smallest distance}} and restrict the derivation to be
 tree-structured.  Co-referential variables can be eliminated by
 replacing them with corresponding antecedent meaning representations.
 
Also note that grammar constrains are enforced by a program validator to ensure that the meaning representations
 are syntactically valid (i.e.,~variables are type-checked)  and
 semantically correct (i.e. no rules logically entail or contradict
 each other).  For example, if the previous meaning representation
 applies a \texttt{Count} rule which returns a number, the next rule
 cannot be \texttt{LookupKey} since the expected argument is a
 database column instead of a number.  If a \texttt{Filter} rule is
 applied to include only restaurants within~500 meters, it does not
 make sense to use a subsequent \texttt{Filter} rule to look for
 restaurants which are more than 1~kilometers away.

 \paragraph*{Step 2: Converting Meaning Representations to Templates} 
 For each meaning representation fragment in the bottom-up derivation,
 we look up the corresponding template underlying the domain-general
 rule that was used to construct it.  The template has missing
 entries, which are instantiated with domain-specific predicates (e.g., \texttt{rel.distance}),
 entities (e.g., \texttt{cuisine.thai}) or referents to previous templates (e.g., \texttt{Result$_{1}$}).  In the end, we obtain a
 sequence of instantiated templates which describe the intention of
 the final meaning representation.

\paragraph*{Step 3: Annotating Utterances}
The instantiated templates are displayed to annotators, who are
asked to create a corresponding utterance.  The collected single-turn utterance does not have to
be a single sentence, it can consist of a few sentences or clauses.
We set no restrictions on the format of expression.  As a result, we
obtain single-turn utterances paired with meaning representations.

\subsection{Data Elicitation for Sequential Utterances 
\label{multi}}

We next discuss how the data elicitation method can be slightly modified to
handle the collection of sequential utterances.  As described in Section~\ref{decom},
sequential utterances  can potentially exhibit a rich class of
co-reference phenomena, either explicit or implicit.  
These phenomena are indicated by co-reference variables in corresponding meaning representations.
We start by analyzing various co-reference structures arising in sequential utterance/meaning representations.
We then proceed to elicit data reflecting these structures.

\begin{figure}[!ht]
	\centering \small
	\begin{tabular}{|@{}l@{\hspace*{-1ex}}cp{4cm}p{5.5cm} |} \hline
		 \multicolumn{4}{|c|}{\bf Exploitation}    \\
		\multirow{3}{*}{\includegraphics[scale=.13]{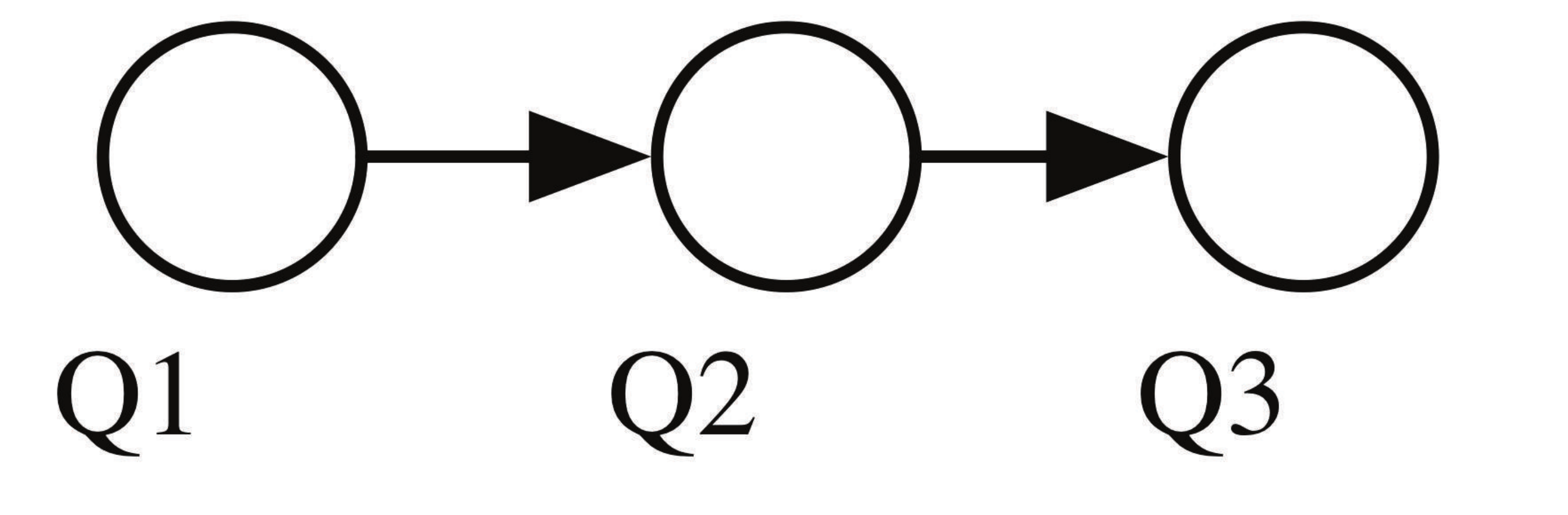}}& Q1: & \emph{restaurants in oxford street?} & \textsl{R$_1$ = find restaurants where location is oxford street} \\
		&   Q2: & \emph{which cost less than 50?} & \textsl{R$_2$ = find [R$_1$] where price $<$ 50} \\
		& Q3: &\emph{with car parking?} & \textsl{R$_3$ = find [R$_2$]which has car parking}\\ \hline
 \multicolumn{4}{|c|}{\textbf{Exploration}}  \\
		\multirow{3}{*}{\includegraphics[scale=.12]{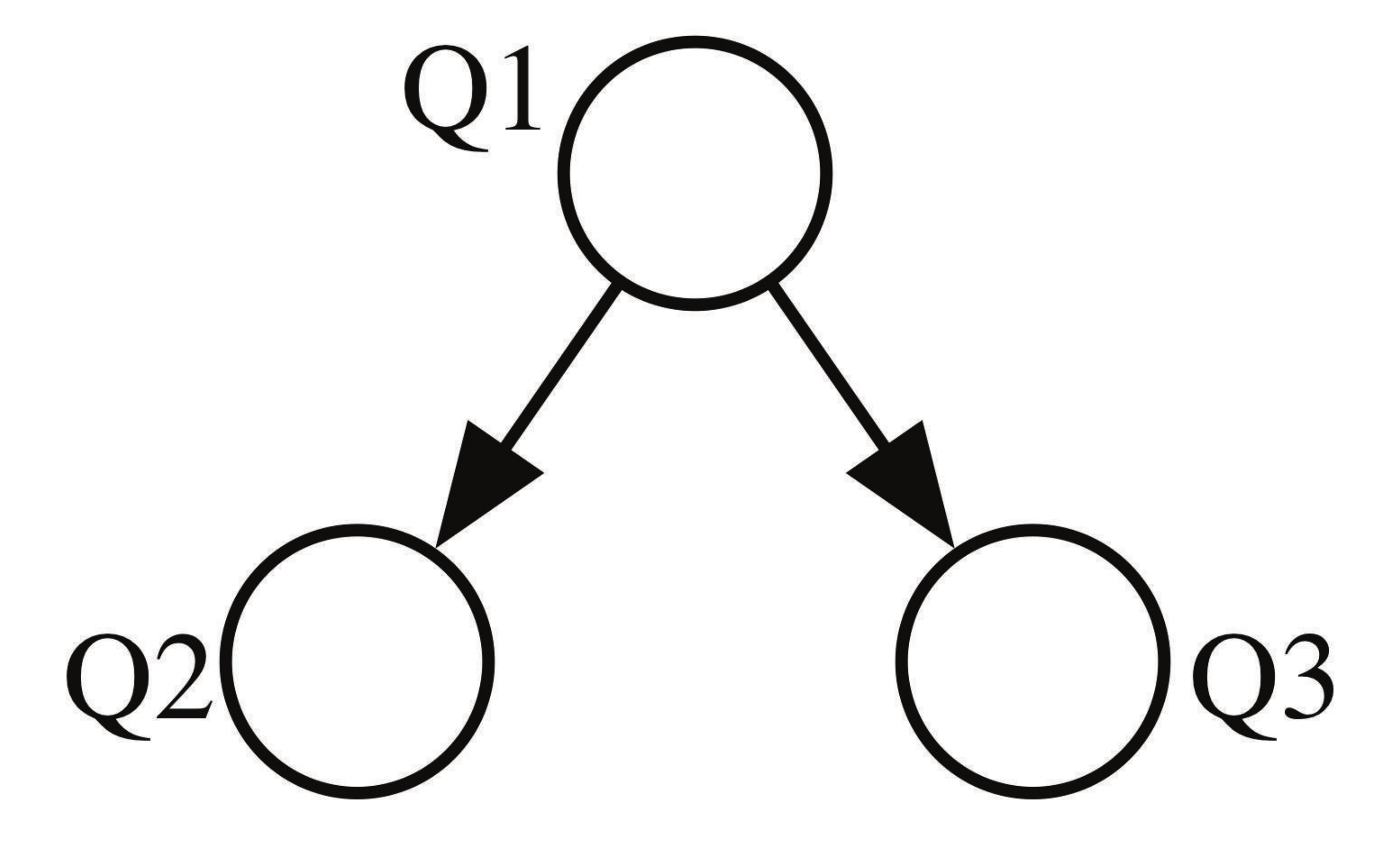}} 
		& Q1: & \emph{restaurants in oxford street} & \textsl{Result$_1$ = find restaurants where location is oxford street}\\
		&   Q2: & \emph{with chinese food?} & \textsl{R$_2$ = find [R$_1$]  where food type is chinese}\\
		& Q3: &\emph{oxford street restaurants with thai food?} & \textsl{R$_3$ = find [R$_1$]  where food type is thai}\\ \hline
		 \multicolumn{4}{|c|}{\bf Merging} \\
		\multirow{3}{*}{\includegraphics[scale=.12]{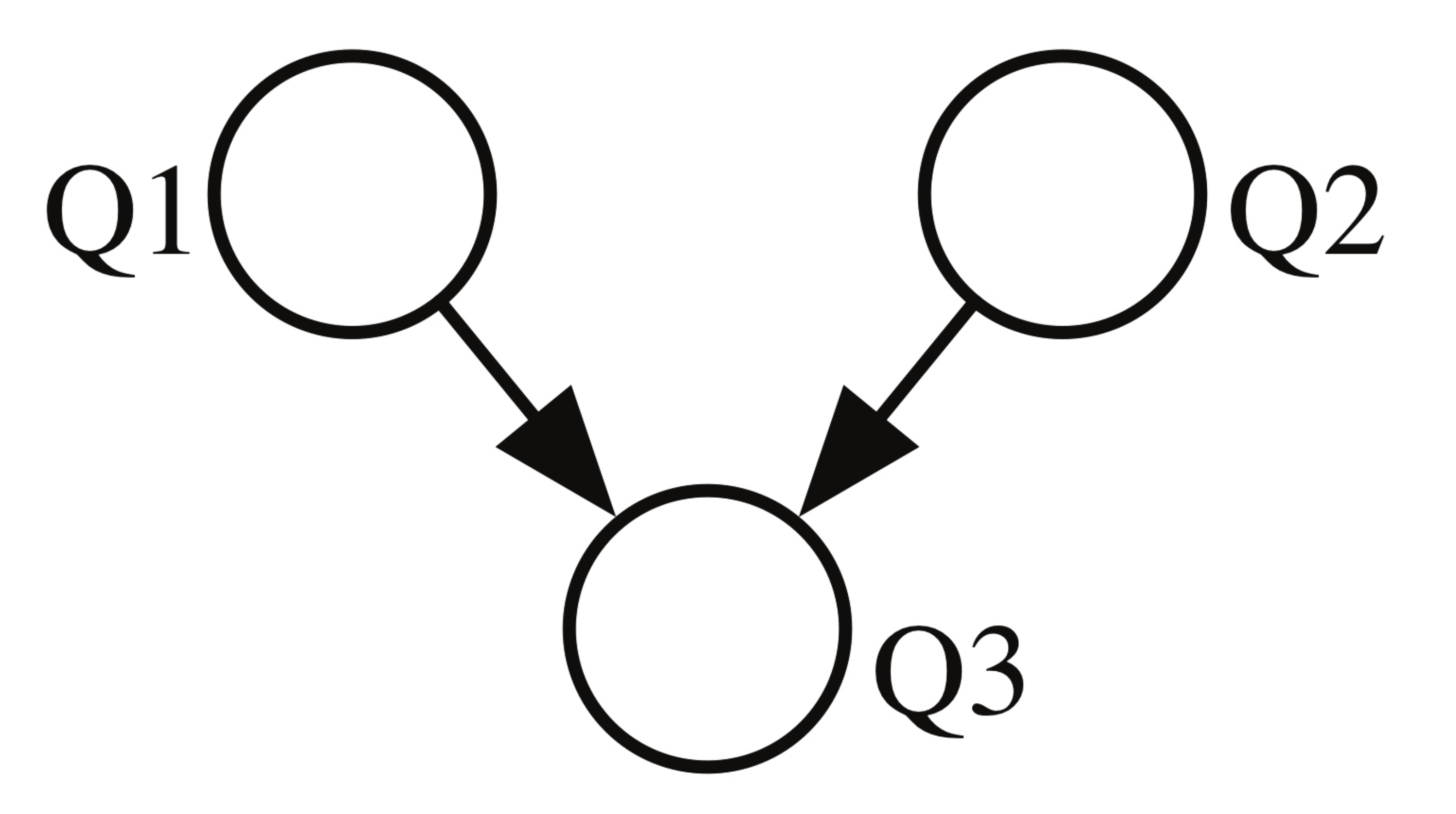}} & Q1:& \emph{find chinese restaurants.} & \textsl{R$_1$ = find restaurants where food type is chinese}\\
		&Q2 & \emph{thai restaurants?} & \textsl{R$_2$ = find restaurants where food type is thai}\\
		& Q3 & \emph{chinese or thai restaurants with car parking?} & \textsl{R$_2$ = find [R$_1$]  and [R$_2$] which has car parking}\\\hline
		\multicolumn{4}{|c|}{\bf Unrelated}  \\
		\multirow{5}{*}{\includegraphics[scale=.12]{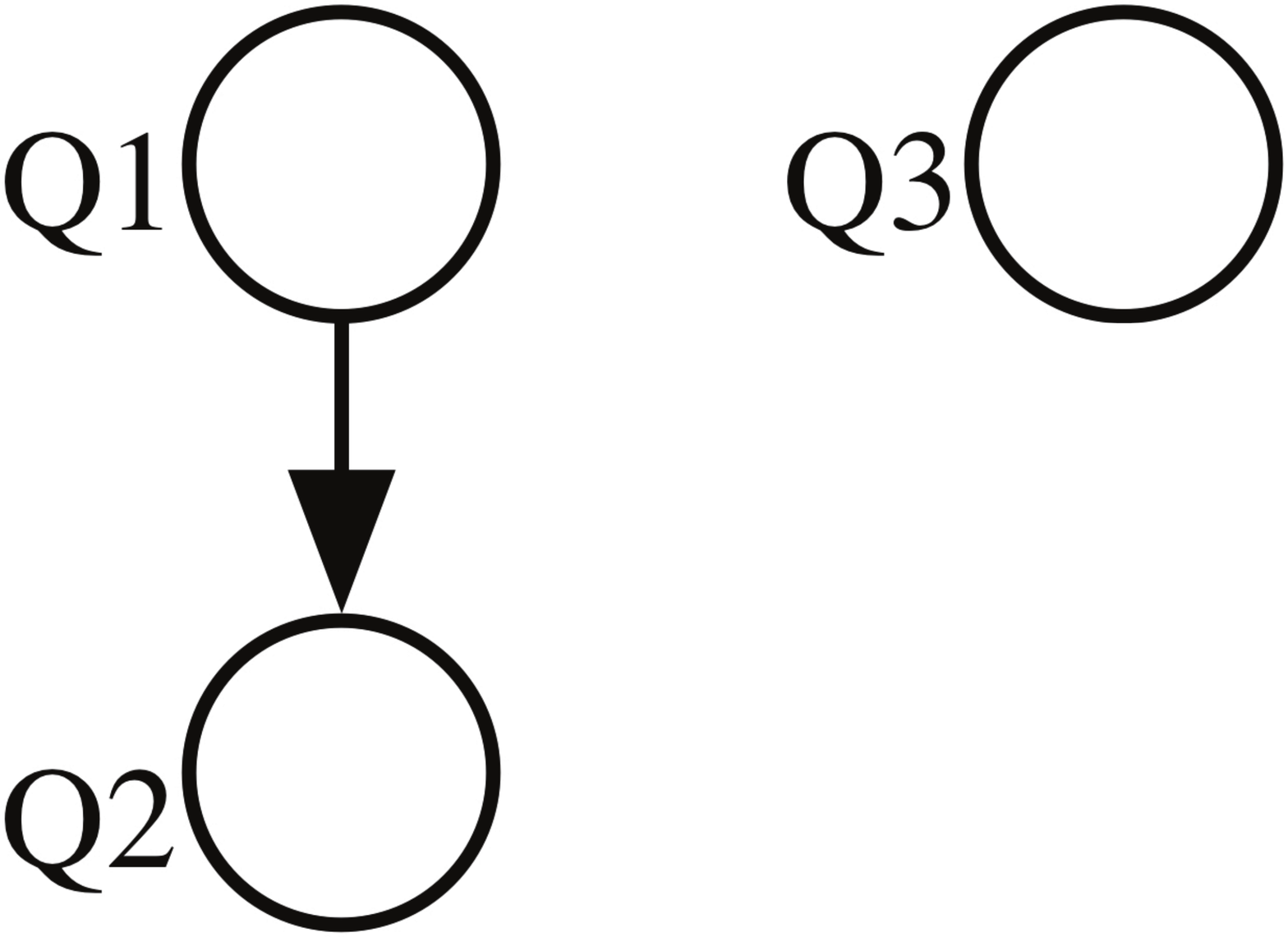}} & Q1: & \emph{which restaurants serve chinese food?} & \textsl{R$_1$ = find restaurants where food type is chinese} \\
		&Q2: & \emph{those in oxford street?} & \textsl{R$_2$ = find [R$_1$]  where location is oxford street}\\
		&Q3: & \emph{which restaurants serve thai food?} & \textsl{R$_3$ = find restaurants where food type is thai} \\
         \hline
	\end{tabular}
	\caption{Coreference structures in three consecutive meaning
          representations and corresponding utterances. They
		are represented as nodes; edges are drawn between
		co-referring utterances; antecedents are nodes with outgoing
		edges while consequents are nodes with incoming edges.}
	\label{coref}
\end{figure}

We follow an ad-hoc, inductive approach which enumerates co-reference
structures in three consecutive meaning representations as a base case,
from which more complex co-reference structures can be constructed.
When the interpretation of meaning representation~M2 (with
corresponding utterance~Q2 and template~R2) depends on meaning
representation~M1 (with corresponding utterance~Q1 and template~R1),
we call M2 (and Q2, R2) the consequent, and M1 (and Q1, R1) the
antecedent.  Co-reference structures in three consecutive utterance/meaning representations are
shown in Figure~\ref{coref} and described in more detail below.

\textbf{Exploitation} refers to chain-structured anaphoric links,
where the consequent of the previous meaning representation is the
antecedent of the next. For example, this happens when a user queries a
database with incremental constrains.  As a result, the next utterance
consistently uses the denotation of the previous one.

\textbf{Exploration} refers to branch-structured anaphoric links,
where an antecedent has two consequents.  For example, this happens
when a user explores different options or constrains related to the same
antecedent.

\textbf{Merging} refers to the inverse of branch-structured anaphoric
links, where a consequent has two antecedents. For example, this
happens when a user combines the denotations of two or more utterances
with union or intersection.  The combined denotation set is then used
in a subsequent utterance.

\textbf{Unrelated} refers to two unconnected co-reference
structures observed in a user session.  For example,
this happens when a user specifies two independent intentions in the same session.

We expect the four categories mentioned above to cover a majority of co-reference structures in sequential utterance/meaning representations. 
However, there exist meanings which they fail to represent or construct. An example is the following sequential utterances:
``\textit{find restaurants in oxford street}'', ``\textit{find restaurants in bond street}'', ``\textit{which of them serve chinese food}'', and 
finally ``\textit{how about those in oxford street}''. 
Note that the purpose of this work is not to formalize all possible co-reference structures. 

In the following, we present our data elicitation method for sequential
utterances which is also illustrated in Table~\ref{mdata example}.

\begin {table}[!ht]
\begin{center}
	\small
\begin{tabular}{|lp{14cm}|} \hline
          1. & Bottom-up construction of meaning representations (done
          by framework) \\ 
          \multicolumn{2}{|c|}{} \\
         & 		Two domain-general rules
          \texttt{$\lambda$s:(lookupKey (var s))}
          and 
          \texttt{$\lambda$s$\lambda$p$\lambda$v:(filter (var s) (var p)
            $=$ (var v))} are instanstiated 
          with  domain-specific variables
          \texttt{type.restaurant},  \texttt{rel.cuisine}, and
          \texttt{cuisine.thai} to obtain the first piece of meaning
          representation: 

\begin{center}
		\texttt{Result$_{1}$=(filter (lookupKey
                  (type.restaurant)) (rel.cuisine) = (cuisine.thai)) }
\end{center}
		
For the second piece of meaning representation,  domain-general rule
\texttt{$\lambda$s:(lookupValue (var s) (var p))} is instantiated with 
domain-specific variables
\texttt{restaurant.kfc} and  \texttt{rel.distance}:  

\begin{center}
\texttt{Result$_{2}$=(lookupValue (restaurant.kfc) (rel.distance))} 
\end{center}

 For the  third piece of meaning representation,  domain-general rule
\texttt{$\lambda$s$\lambda$p$\lambda$v:((var s) min (var p))} is
instantiated  with domain-specific variables
\texttt{Result$_{1}$} and \texttt{rel.price}: 

\begin{center}
\texttt{Result$_{3}$=((Result$_{1}$) argmin (rel.price))}
\end{center}

The final piece of meaning representation is constructed with the
domain-general rule \texttt{$\lambda$s$\lambda$p$\lambda$v:(filter
  (var s) (var p) $<$ (var v))} and domain-specific variables
\hspace*{.34cm}\texttt{Result$_{3}$}, \texttt{rel.distance}, and
\texttt{Result$_{2}$}: 
\begin{center}
\texttt{Result$_{4}$=(filter (Result$_{3}$) (rel.distance) $<$ (Result$_{2}$))}\\
\end{center} \\ \hline 

2. & Each piece of meaning representation is
converted to a canonical representation described by templates.
Templates associated with domain-general rules are instantiated with
domain-specific and co-referential variables (shown in brackets): \\
\multicolumn{2}{|c|}{}\\

& \textit{Result$_{1}$ = find} [\textit{restaurants}] \textit{where} [\textit{cuisine}] \textit{is} [\textit{Thai}] \\

&\textit{Result$_{2}$ = find}  [\textit{distance}] of [\textit{KFC}]\\

& \textit{Result$_{3}$ = find} [\textit{Result$_{1}$ with smallest }] [\textit{price rating}]\\

& \textit{Result$_{4}$ = find} [\textit{Result$_{3}$}] \textit{with} [\textit{distance}] \textit{$<$} [Result$_{2}$] \\\hline
		
3. & The templates are displayed to  annotators who summarize them
into an utterance:
\begin{center}
		\textit{Show me Thai restaurants.} \\
		\textit{How far is KFC?}\\
		\textit{Which  Thai restaurants are cheapest?} \\
		\textit{Which of these  are closer to me than KFC?}\\
\end{center}\\\hline
	\end{tabular}
\end{center}
\caption{Example of our data elicitation procedures for sequential
  utterances paired with meaning representations.}  
\label{mdata example}
\end{table}

\paragraph*{Step1: Generating Meaning Representations}

As in the single-turn case, we use a context-free grammar and a computer program to generate
meaning representations by sampling domain-general and specific rules.
Again, the generation is performed in a bottom-up manner, so that
larger pieces of meaning representation can be constructed from
smaller ones.  Each domain-general rule is instantiated with
domain-specific or co-referential variables and results in a new piece
of meaning.  Different from the single-turn case, we allow each
meaning representation to be constructed with one to three
domain-general rules, so as to have some basic level of
compositionality.  Furthermore, we consider a richer class of
co-reference structures in-between meaning representations (in the
single-turn case, the co-reference between meaning representations is
mostly chain-structured).

We use the three terminal-level rules introduced in Table~\ref{coreft} to generate
 co-referential variables.  For each co-referential
variable, we additionally generate its value---which is selected from the index list
of previously generated meaning representations.  To recap, we assign an index to each meaning representation
(e.g.,~\texttt{Result$_{1}$}, \texttt{Result$_{2}$}) and this index is
used to denote the value of a co-referential variable.  Similar to
the single-turn case, grammar constrains are enforced by a validator to ensure the meaning representations are
syntactically and semantically valid (e.g., if the previous meaning
representation applies a \texttt{Count} rule which returns a number,
the next rule cannot be \texttt{LookupKey} since the expected argument
is a database column instead of a number).  We use additional
constrains to guarantee that the anaphoric links are valid; we cache
the~$n$ most recently generated meaning representations, their
consequents and antecedents, and return types.  For the next meaning
representation, we sample terminal-level rules which generate
co-referential variables and their values, by sampling one of the cases
below:
\begin{enumerate}
	\itemsep0em
      \item the co-referential variable is \texttt{Coref}, and its
        value is one of the previously generated meaning
        representations whose denotation is type checked and has no
        consequents.  This results in the Exploitation structure.
      \item the co-referential variable is \texttt{Coref}, and its
        value is one of the previously generated meaning
        representations whose denotation is type checked and has other
        consequents.  This results in the Exploration structure.
      \item the co-referential variable is \texttt{Union\_coref} or
        \texttt{Intersection\_coref}, and its value is constructed
        from two previous meaning representations whose denotations
        are type checked. Moreover, the two meaning representations
        should not exhibit an Exploitation structure.  This results in
        the Merging structure.
      \item there is no co-referential variable in the meaning
        representation. This results in the Unrelated structure.
\end{enumerate}

\paragraph*{Step 2: Converting Meaning Representations to Templates} 
For each meaning
representation in the sequence, we retrieve the domain-general rule it
uses, and look up the corresponding template.  The template has
missing entries, which are instantiated with domain-specific
predicates, entities, or co-referential index.

Since meaning representations may be constructed with more than one
domain-general rules, there may be more than one initial templates associated
with each meaning representation. We combine these templates by
merging their constrains.  For example, the two templates
`\textit{Result$_{1}$ = find} [\textit{restaurants}] \emph{with}
[\emph{distance}] $<$ [\emph{500m}]'' and \textit{Result$_{2}$ = find}
[\textit{Result$_{1}$}] \emph{with} [\emph{price}] $>$ [\emph{50\$}]
are merged into \textit{Result$_{1}$ = find} [\textit{restaurants}]
\emph{with} [\emph{distance}] $<$ [\emph{500m}] and [\emph{price}] $>$
[\emph{50\$}].  This ensures that every meaning representation has
exactly one canonical template representation.

\paragraph*{Step 3: Annotating Utterances} 
We display the sequence of instantiated templates to crowdworkers.
Different from the single-turn case where crowdworkers summarize all
the templates into a single utterance, in the sequential case we ask
them to paraphrase every template into an utterance, while expressing
the discourse structure of the whole task (by using implicit or
explicit co-reference).  This results in a sequence of inter-related
utterances paired with meaning representations.

\section{Neural Semantic Parsing \label{nsp}}

\begin{figure}[t]
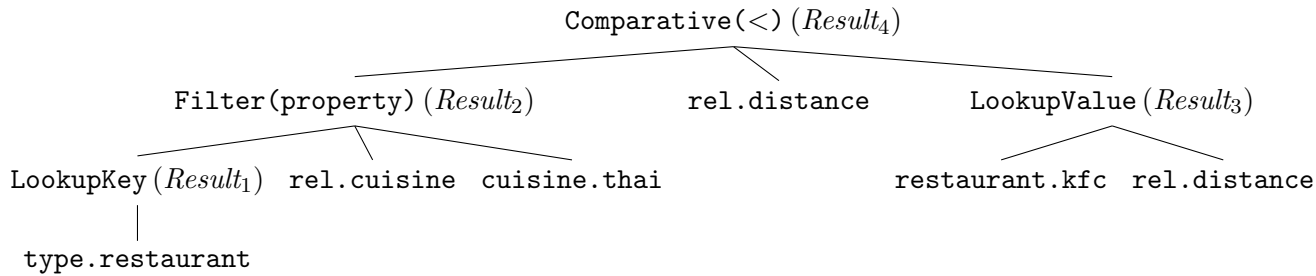

	\centering
		\Tree[.\texttt{Comparative($<$)}\,(\textit{Result$_4$})
		        [.\texttt{Filter(property)}\,(\textit{Result$_2$}) 
		        [.\texttt{LookupKey}\,(\textit{Result$_1$}) \texttt{type.restaurant} ] \texttt{rel.cuisine} \texttt{cuisine.thai} ]
		        \texttt{rel.distance} [.\texttt{LookupValue}\,(\textit{Result$_3$}) \texttt{restaurant.kfc} \texttt{rel.distance} ] ]
		\caption{Derivation tree for meaning representation in
                  Table~\ref{tab:single}. Domain-general rules are
                  represented as non-terminal nodes, using the
                  abbreviations shown in
                  Table~\ref{general-aspects}. For example,
                  \texttt{LookupKey} refers to
                  \texttt{$\lambda$s:(lookupKey (var
                    s))}. Domain-specific and co-referential rules or
                  variables are represent as terminal nodes.
                        \label{tree}}
\end{figure}

A feature of our data elicitation method is that it explicitly
models the compositional process underlying meaning representations,
and caches the rules applied recursively to derive them.  This
characteristic allows us to train a neural semantic parser which
generates meaning representations by predicting and composing rules
therein.  Figure~\ref{tree} shows an example of the derivation our
model predicts.

\subsection{Transition Algorithm for Generating Derivations}
We introduce an algorithm for
generating tree-structured derivations with three classes of transition actions.  The
key insight underlying our algorithm is to define a canonical generation order, which produces a tree as a sequence of configuration-transition pairs
[$(c_0, t_0), (c_1, t_1), \cdots, (c_m, t_m)$]. 
In this work, we use top-down order for generation.

The top-down system is specified by tuple $c=(\sum, \pi, \sigma, N,
P)$ where $\sum$ is a stack used to store partially complete tree
fragments, $\pi$ denotes non-terminal rules to be generated in the
derivation tree, $\sigma$ denotes terminal rules to be generated, $N$
is a stack of incomplete non-terminal rules, and $P$ is a function
indexing the position of a non-terminal pointer. The pointer indicates
where subsequent rules should be attached (e.g.,~$P(X)$ means that the
pointer is pointing to the non-terminal $X$ and the next rule is
generated underneath $X$).  We take the initial configuration of the
transition system to be $c_0 = ([], TOP, \varepsilon, [], \bot)$,
where $TOP$ stands for the root node of the tree, $\varepsilon$
represents an empty string, and $\bot$ represents an unspecified
function. The top-down system employs three classes of transition
operations defined in Table~\ref{td_formal}:

\begin {table}[t]
\begin{center}
	\begin{tabular}{l l }
		\hline
		\multicolumn{2}{c}{\textbf{Top-down Transitions}} \\\hline
		\texttt{NT(X)}  &  $([\sigma | \texttt{X}'], \texttt{X}, \varepsilon,  [\beta | \texttt{X}'], P(\texttt{X}')) \Rightarrow  ([\sigma | \texttt{X}', \texttt{X}], \varepsilon, \varepsilon, [\beta | \texttt{X}', \texttt{X}], P(\texttt{X})) $  \\
		\texttt{TER(x)}  & $([\sigma | \texttt{X}'], \varepsilon, \texttt{x}, [\beta | \texttt{X}'], \texttt{P(X')}) \Rightarrow  ( [\sigma | \texttt{X}', \texttt{x}], \varepsilon, \varepsilon , [\beta | \texttt{X}', \texttt{x}], P(\texttt{X}'))  $  \\
		\texttt{RED} & $([\sigma | \texttt{X}', \texttt{X}, \texttt{x}], \varepsilon, \varepsilon, [\beta | \texttt{X}', \texttt{X}], P(\texttt{X})) \Rightarrow  ([\sigma | \texttt{X}', \texttt{X}(\texttt{x})], \varepsilon, \varepsilon, [\beta | \texttt{X}'], P(\texttt{X}')) $ \\
		\hline
	\end{tabular}
\end{center}
\caption{Transition actions for the top-down generation system. Stack~$\sum$
	is represented as a list with its head to the right (with tail
	$\sigma$). \label{td_formal}} 
\end{table}

\begin{itemize}
	\item \texttt{NT(X)} creates a new subtree rooted by a non-terminal rule \texttt{X}.  The non-terminal \texttt{X} is pushed on top of the
	stack and written as \texttt{X(}.  The token \texttt{(} implies $X$ is incomplete and subsequent rules are
	generated as children underneath~\texttt{X}. In our semantic formalism, \texttt{X} can be any domain-general rule in~Table \ref{general-aspects}.
	\item \texttt{TER(x)} creates a new terminal rule \texttt{x}. The terminal \texttt{x} is pushed on top
	of the stack, written as \texttt{x}. In our semantic formalism, \texttt{x} is one category of the
	domain-specific or co-referential rules shown
	in the first column of~Table \ref{specific-rule} and Table \ref{coreft}. At this stage, \texttt{TER(x)} simply generates a variable (i.e., a category) whose value is yet to be 
	determined by a domain-specific classifier. 
	\item \texttt{RED} is the reduce operation which indicates that the
	current subtree being generated is complete. 
	The non-terminal rule 
	of the current subtree will be applied to the children rules underneath (i.e. a beta reduction in lambda calculus), 
	formulating a large piece of meaning representation.
	This  large piece of meaning representation serves as a single child to its predecessor non-terminal,
	and subsequent children rules will
	be attached to the predecessor.  Stack-wise,
	\texttt{RED} recursively pops children rules (which can be either
	terminal rules or partial meaning representations) on top until an incomplete non-terminal rule
	is encountered.  The non-terminal is popped as well and beta-reduced, after which 
	the larger piece of meaning representation is pushed back to the stack as a single closed
	constituent, written for example as \texttt{X1}(\texttt{X2},
	\texttt{X3}).
\end{itemize}

The space of transition actions is domain-general and language dependent. It covers
operations that generate non-terminal-level rules (i.e., domain-general rules),
terminal-level variables (for domain-specific and co-referential rules),
and \texttt{RED} which enforces tree structures.  Notice
that \texttt{TER(x)} generates terminal-level variables only but not
their values. This makes the entire generation algorithm portable
across domains or transferable to new domains. Specific values for terminal-level variables
(e.g.,~which domain-specific rule should be used, or which utterance
the anaphora links to) will be predicted by subsequent classifiers, as
an instance of hierarchical classification.  The sequence of
transition actions which generate the tree in Figure~\ref{tree} is
as follows:
\begin{quote}
\texttt{NT(Comparative($<$))},  \texttt{NT(Filter(property))},  \texttt{NT(LookupKey)},   \texttt{TER(Entity)}, \texttt{RED}, 
\texttt{TER(Binary\_predicate)},  \texttt{TER(Entity)}, \texttt{RED},  \texttt{TER(Binary\_predicate)},  \texttt{RED},  \texttt{NT(LookupValue)},  
\texttt{TER(Entity)}, \texttt{TER(Binary\_predicate)}, \texttt{RED},
\texttt{RED}.
\end{quote}

\subsection{Neural Network Realizer \label{nn}}
We model the above generation algorithm with a neural sequence-to-tree
model, which encodes the utterance and then predicts a sequence of
transition actions which generate rules to construct the
derivation tree of the meaning representation.

\paragraph{Encoder} Each utterance~$x$ is encoded with a bidirectional
LSTM \cite{hochreiter1997long}.  A bidirectional LSTM is
comprised of a forward LSTM and a backward LSTM.  The forward LSTM
processes a variable-length sequence $x=(x_1, x_2, \cdots, x_n)$ by
incrementally adding new content into a single memory slot, with gates
controlling the extent to which new content should be memorized, old
content should be erased, and current content should be exposed. At
time step~$t$, the memory~$\vec{c_t}$ and the hidden state~$\vec{h_t}$
are updated with the following equations:
\begin{equation}
\begin{bmatrix}
i_t\\ f_t\\ o_t\\ \hat{c}_t
\end{bmatrix} =
\begin{bmatrix} \sigma\\ \sigma\\ \sigma\\ \tanh
\end{bmatrix} W\cdot [\vec{h_{t-1}}, \, x_t]
\label{beginlstm}
\end{equation}
\begin{equation} \vec{c_t} = f_t \odot \vec{c_{t-1}} +
i_t \odot \hat{c}_t
\end{equation}
\begin{equation} \vec{h_t} = o_t \odot \tanh(\vec{c_t})
\label{endlstm}
\end{equation} 
where $i$, $f$, and $o$ are gate activations; $W$ denotes the weight matrix. For simplicity, we denote the recurrent computation of the forward LSTM as:
\begin{equation} \vec{h_t} = \vec{\textnormal{LSTM}} (x_t, \vec{h_{t-1}})
\label{lstm1}
\end{equation} 
After encoding, a list of token representations $[\vec{h_1},
\vec{h_2}, \cdots, \vec{h_n}]$ within the forward context is obtained.
Similarly, the backward LSTM computes a list of token representations 
$[\cev{h_1}, \cev{h_2}, \cdots, \cev{h_n}]$ within the backward context as:
\begin{equation} \cev{h_t} = \cev{\textnormal{LSTM}} (x_t, \cev{h_{t+1}})
\label{lstm2}
\end{equation} 

Finally, each input token $x_i$ is represented by the concatenation of
its forward and backward LSTM state vectors, denoted by $h_i =
\vec{h_i} : \cev{h_i}$.  The list storing token vectors for the entire
utterance $x$ is denoted by $b = [h_1, \cdots, h_k]$, where $k$ is the
length of the utterance.

\paragraph{Decoder}
After the utterance is encoded, the derivation tree of the
corresponding meaning representation is generated with a stack-LSTM
decoder, whose state updates depend on the class of transition
operations.  Specifically, transition actions \texttt{NT} and
\texttt{TER} change the stack-LSTM representation~$s_t$ as in a
vanilla LSTM:
\begin{equation}
s_t = \textnormal{LSTM} (y_t, s_{t-1})
\end{equation}
where~$y_t$ denotes the newly generated non-terminal or terminal
rule. 

Transition action \texttt{RED} recursively pops the stack-LSTM
states as well as corresponding tree nodes on the output stack. The
popping stops when a state for a non-terminal rule is reached and
popped, after which the stack-LSTM reaches an intermediate state
$s_{t-1:t}$.\footnote{We use $s_{t-1:t}$ to denote the interim state
  from time step~$t-1$ to~$t$, after popping; $s_t$ denotes the final
  LSTM state after the subtree representation is pushed back to the
  stack (as explained in the following).} The representation of the
completed subtree~$u$, which represents a large piece of meaning
representation, is then computed as:
\begin{equation}
u = W_u \cdot [p_u : c_u]
\end{equation}
where $p_u$ denotes the non-terminal rule in the subtree, $c_u$
denotes the average of the children embeddings underneath, and $W_u$
denotes the weight matrix.  Note that $c_u$ can also be computed with
more advanced method such as a recurrent neural network
\cite{dyer2016recurrent}.  Finally, the subtree embedding~$u$ serves
as the input to the LSTM and updates $s_{t-1:t}$ to $s_t$ as:
\begin{equation}
s_t = \textnormal{LSTM} (u, s_{t-1:t})
\label{reduce:1}
\end{equation}

\paragraph{Predictions for Transition Actions}
At each time step~$t$, the decoder needs to predict the next
transition action.  This prediction is 
conditioned on the utterance (represented by the bidirectional-LSTM
states $b = [h_1, \cdots, h_k]$), and the partially completed
derivation tree (represented by the current state of the stack-LSTM
$s_t$).  To make the prediction, we compute an adaptive representation
of the utterance~$\bar{b}_t$ with a soft attention mechanism:
\begin{equation}
u_t^i  = V \tanh (W_b b_i + W_s s_t) 
\label{att1}
\end{equation}
\begin{equation}
\alpha_t^i  =  \textnormal{softmax} (u_t^i )
\end{equation}
\begin{equation}
\bar{b}_t  = \sum_i  \alpha_t^i  b_i 
\label{att3}
\end{equation}
where $W_b$ and $W_s$ are weight matrices and $V$ is a weight vector.
We then combine~$\bar{b}_t$ and  $s_t$ with a
feed-forward neural network (Equation~(\ref{softmax})) to yield a
feature vector for the generation system. Finally,
$\textnormal{softmax}$ is taken to obtain the parameters of the
multinomial distribution over actions:
\begin{equation}
a_{t+1}  \sim \textnormal{softmax} (  W_{oa} \tanh( W_f [\bar{b}_t, s_t] )  )
\label{softmax}
\end{equation}
where $W_{oa}$ and $W_f$ are weight matrices.
In greedy decoding, $a_{t+1}$ is selected as the action which has the highest probability.

In cases where transition action  \texttt{TER(x)} generates variables of
predicates, entities, or co-reference, we need to further predict
their values. 

\paragraph{Predictions for Variable Values} 

For predicates and entities, we use a soft attention layer similar to
Equations~(\ref{att1})--(\ref{att3})  to compute the adaptive
utterance representation~$\bar{b}_t$; and then predict specific
entities or predicates with another $\textnormal{softmax}$ 
classifier as:
\begin{equation}
y_{t+1}  \sim \textnormal{softmax} (  W_{oy} \tanh( W_f [\bar{b}_t, s_t] )  )
\end{equation}
which outputs multinomial distribution over predicates or entities,
and this set of neural parameters is domain-specific.


To predict the value of co-referential variables, we
rely on a self-attention network
\cite{cheng2016intra,parikh2016decomposable} to determine which
utterances it co-refers with. The number of predictions is determined
by the type of the co-referential variable.  For variable
\texttt{Coref}, we need to predict one anaphoric link, whereas for
\texttt{Union\_coref} and \texttt{Intersection\_coref} we need to
predict two.

Given utterance~$x$, our previous bidirectional LSTM encoder obtains a
list of forward representations $[\overrightarrow{h_1}, \cdots,
\overrightarrow{h_n}]$ and backward representations
$[\overleftarrow{h_n}, \cdots, \overleftarrow{h_1}]$.  We use the
concatenation of $\overrightarrow{h_n}$ and $ \overleftarrow{h_1}$ as
utterance representation $X$, which is fed to the self-attention
network. Given the current utterance embedding $X_c$ and the previous
utterance embeddings $X_1, \cdots, X_{c-1}$, the self-attention
network computes the relation distribution between~$X_c$ and every
previous utterance in the list of [$X_1, \cdots, X_{c-1}$].  This is
accomplished by first computing score~$u_c^i$ for each previous
utterance in the list, with index~$i$ ranging from~$1$ to~$c-1$:
\begin{equation}
u_c^i  = W_v \tanh (W_{i} X_i + W_{X} X_c) 
\label{corefeq}
\end{equation}
Then a $\textnormal{softmax}$ classifier is used to obtain~$a_c^i$,
the probability that the current utterance $X_c$ co-refers with $i$th
utterance $X_i$:
\begin{equation}
a_c^i =  \textnormal{softmax} (u_c^i )
\end{equation}
where the $\textnormal{softmax}$ is taken over all previous utterances
in the list, and $W$s are weight parameters.  We then select the most
probable utterance as the value of \texttt{Coref}, or the top two most probable
utterances as the value of \texttt{Union\_coref} or
\texttt{Intersection\_coref}.

\subsection{Context-dependent  Parsing (Optional)} 
The neural semantic parser presented above is able to generate meaning
representations with and without co-reference, and thus handle both
single-turn and sequential utterances.  However, it should be noted
that when parsing sequential utterances, our semantic parser does not
make an explicit use of context: the generation of a co-referential
variable relies on the current utterance only, before the variable's
value is predicted within context.  In the following, we extend our model
to make an explicit use of context when parsing sequential
utterances.

The central idea is to maintain a context vector for the current user
session.  This vector is used as an additional feature to the softmax
classifier (Equation~\eqref{softmax}) that predicts transition
operations.  To compute the context vector, we use the same
self-attention network described in Equation~\eqref{corefeq}.  Given
current utterance $x_c$ with representation $X_c$, the network
computes score $u_c^i$ for each of the previous utterances $x_i$, with
representation~$X_i$ and index~$i$ ranging from~$1$ to~$c-1$:
\begin{equation}
u_c^i  = W_v \tanh (W_{i} X_i + W_{X} X_c) 
\end{equation}
We then compute a context representation for the current utterance,
denoted by $\bar{X}_c$:
\begin{equation}
a_c^i =  \textnormal{softmax} (u_c^i )
\end{equation}
\begin{equation}
\bar{X}_c = \sum a_c^i  * X_c
\end{equation}
The context representation~$\bar{X}_c$ is used at every time step when
the decoder generates the derivation tree of $x_c$, by extending
Equation~\eqref{softmax} as:
\begin{equation}
a_{t+1}  \sim \textnormal{softmax} (  W_{oa} \tanh( W_f [\bar{b}_t, s_t, \bar{X}_c] )  )
\end{equation}
We will experimentally verify whether modeling context explicitly is
useful for parsing sequential utterances.

\section{Experiments \label{exp}}
In this section, we describe a range of experiments conducted to
evaluate the framework which builds neural semantic parsers from domain ontologies.  Specifically, we will present experimental results and analysis for both single-turn
and sequential utterances.

\subsection{Data Elicitation}
\label{sec:data-collection}

\subsubsection{Single-turn Utterances}
We elicited single-turn utterances for six domains simulating database
querying tasks. Two domains relate to company management (meeting and
employees databases), two concern recommendation engines (hotel and
restaurant databases), and two target healthcare applications (disease
and medication databases).

The dataset was collected with Amazon Mechanical Turk (AMT).  Across
domains, the total number of querying tasks (described by templates)
that the framework generated was~7,225.  These tasks were sampled
randomly without replacement to show to annotators. Each annotator
saw three tasks per HIT and was paid~0.3\$.  After removing repeated
query-meaning representation pairs, we collected a semantic parsing dataset of
7,708 examples.  The average amount of time annotators spent on each
domain was three hours.  We evaluated the correctness of 100 randomly
chosen utterances, and the accuracy was 81\%.  
For the failed examples, annotators generated utterances constitute partial or wrong descriptions of the meaning representations.


\begin{table*}[t]
	\begin{center}
		\small
		\begin{tabular}{|@{~}l@{~}|@{~}r@{\hspace{.3ex}}r@{\hspace{.5ex}}r@{\hspace{.5ex}}r@{~} |@{~} r@{\hspace{.5ex}}r@{\hspace{.5ex}}r@{\hspace{.5ex}}r@{~} |@{~}r@{\hspace{.5ex}}r@{\hspace{.5ex}}r@{~}r | @{~}r@{\hspace{.5ex}}r@{\hspace{.5ex}}r@{\hspace{.5ex}}r@{~}|@{~}r@{\hspace{.5ex}}r@{\hspace{.5ex}}r@{\hspace{.5ex}}r@{~}|}
			\hline
			\#rules & \multicolumn{4}{c@{~}|@{~}}{1} & \multicolumn{4}{c@{~}|@{~}}{2} & \multicolumn{4}{c@{~}|@{~}}{3} &  \multicolumn{4}{c@{~}|@{~}}{4} & \multicolumn{4}{c|}{All} \\ 
			Domain & \multicolumn{1}{@{~}c@{~}}{Q} & \multicolumn{1}{@{~}c@{~}}{Tp} & \multicolumn{1}{@{~}c@{~}}{Tk}  & \multicolumn{1}{@{~}c|@{~}}{WO}& \multicolumn{1}{c@{~}}{Q} & \multicolumn{1}{@{~}c@{~}}{Tp} & \multicolumn{1}{@{~}c@{~}}{Tk } & \multicolumn{1}{@{~}c|@{~}}{WO}& \multicolumn{1}{@{~}c@{~}}{Q} & \multicolumn{1}{@{~}c@{~}}{Tp} & \multicolumn{1}{@{~}c@{~}}{Tk}  & \multicolumn{1}{@{~}c|@{~}}{WO}& \multicolumn{1}{@{~}c@{~}}{Q} & \multicolumn{1}{@{~}c@{~}}{Tp} & \multicolumn{1}{@{~}c@{~}}{Tk}  & \multicolumn{1}{@{~}c|@{~}}{WO}& \multicolumn{1}{@{~}c@{~}}{Q} & \multicolumn{1}{@{~}c}{Tp} & \multicolumn{1}{@{~}c@{~}}{Tk} &\multicolumn{1}{@{~}c@{~}|}{WO}\\\hline
			meeting   & 378 & 191&	9.1 & 1.41  & 570 &537 &16.20 &2.21    &254 &254 &16.2 &2.81  &46  &46 &21.37 & 3.19  &	1,248 & 1,028 & 12.59 & 2.21 \\
			employees  & 322 & 240&	8.96 & 1.34 & 320 &319 &13.47 &2.95    &486 &486 &18.2 &3.66  &268 &268 &22.37    & 3.12  &1,396 & 1,313 & 15.79 & 3.12\\
			hotel	  &170  & 146&	8.99& 1.91 & 358  &542 &16.86 &3.64    &542 &542 &16.86 &4.11  &433 &433 &19.89    & 4.05  &1,503 & 1,479 & 15.87 & 4.05\\
			restaurant&132  &  98&	8.14& 1.40 & 301  &295 &12.74 &3.41    &495 &495 & 16.74 & 3.74 &311 &311&20.02    & 3.66  &1,239 & 1,199 & 15.68 & 3.66\\
			disease	  &283  &212&	9.3 & 1.29 & 301  &455 &14.23 &3.52    &455 &455 & 19.49 & 5.65 & 213 &213& 23.95& 4.48 &	1,252 & 1,176 & 16.68 & 4.48\\
			medication&136	&102 &8.53 & 1.31 & 252  &246  &11.89 &2.35    &435 &435 & 16.09 &3.17  &247 &247&20.04	& 2.91 & 1,070 & 1,030 & 15.05 & 2.91\\\hline
			
		\end{tabular}
	\end{center}
	\caption{\textsc{SingleTurn} dataset: number of utterances (Q), number of templates (Tp), average
          number of tokens (Tk) and token overlap between queries and
          templates (WO) per domain and overall (All). \label{nqueries}}
\end{table*}

Table~\ref{nqueries} shows various statistics of the elicited dataset
which we call \textsc{SingleTurn}. These include the number of
utterances (Q) we obtained for each domain broken down according to
the depth of compositionality, the number of templates (Tp) per
domain, the average number of tokens (Tk) per utterance in each domain,
and the token overlap (Ov) between the utteranes and the corresponding
templates.  The number of utterances collected at
each compositional level is dependent on the space of predicates in
each domain.  We  see that utterance length does not vary
drastically among domains even though the average number of tokens is
affected by the verbosity of entities and predicates in each
domain. We use word overlap as a measure of the amount of
paraphrasing.  We see that utterances in the disease domain deviate
least from their corresponding templates while utterances in the
meeting domain deviate most.  This number is affected by the degree of
expert knowledge required for annotation in each domain.

Table~\ref{dataexamples} presents examples of the utterances we elicited
for the six domains together with their corresponding templates.

\begin{table}[!ht]
	\begin{center}
		\scriptsize
		\begin{tabular}{|@{~}l |p{6cm} | p{6cm} @{~}|} \hline
			\multicolumn{1}{|c|}{Domain}    &   \multicolumn{1}{c|}{Templates} &  \multicolumn{1}{c|}{Query}   \\ \hline
			meeting   &
			\textit{R$_{1}$ = find}
			[\textit{location}]  \textit{of} [\textit{annual review}]& \textit{what location will the annual review take place}\\ \hline
			meeting   &
			\textit{R$_{1}$ = find}
			[\textit{location}]  \textit{of} [\textit{annual review}] \newline 
			\textit{R$_{2}$ = find all}
			[\textit{meetings}] \newline
			\textit{R$_{3}$ = find}
			[\textit{R$_{2}$}]  \textit{where} [\textit{location}] is not [\textit{R$_{1}$}] \newline 
			\textit{R$_{4}$ = find}
			[\textit{R$_{3}$}]  \textit{with smallest number of} [\textit{attendee}]
			& \textit{which meeting is not held in the same venue as annual review, and attracts the least amount of attendance}\\ \hline                 
			
			employees 
			& \textit{R$_{1}$ =
				find all}
			[\textit{employees}] \newline
			\textit{R$_{2}$ = find}
			[\textit{R$_{1}$}]
			\textit{with  smallest number of} [\textit{projects}] &  \textit{which employee is assigned minimum projects} \\ \hline
			
			employees 
			& \textit{R$_{1}$ =
				find all}
			[\textit{employees}] \newline
			\textit{R$_{2}$ = find}
			[\textit{R$_{1}$}]
			\textit{with} [\textit{salary}] \textit{$<$} [5000]  \newline
			\textit{R$_{3}$ = find}
			[\textit{R$_{1}$}]
			\textit{with} [\textit{salary}] \textit{$>$} [15000]  \newline
			\textit{R$_{4}$ = find}
			[\textit{R$_{2}$ or R$_{3}$}]
			\textit{where} [\textit{division}] \textit{is} [IT]  
			&  \textit{for those employees in the IT division, who are paid less than 5000 or more than 15000} \\ \hline
			
			hotel     &
			\textit{R$_1$ = find all}  [\textit{hotels}] \newline \textit{R$_2$ = find} [\textit{R$_1$}] \textit{where} [\textit{distance}] is  [\textit{300}] \newline \textit{R$_3$ = find} [\textit{R$_2$}] \textit{which satisfies} [\textit{free cancellation}]& \textit{which hotel is at 300 metres and doesn't charge a cancellation fee} \\ \hline
			hotel     &
			\textit{R$_1$ = find all}  [\textit{hotels}] \newline 
			\textit{R$_2$ =  find} [\textit{R$_1$}] \textit{where} [\textit{location}] \textit{is} [\textit{oxford street}] \newline
			\textit{R$_3$ =  find} [\textit{R$_2$}] \textit{where} [\textit{room type}] \textit{is} [\textit{single} or \textit{double}] \newline
			\textit{R$_4$ =  find} [\textit{R$_3$}] \textit{which satisfies} [\textit{has free wifi}]
			& 
			\textit{which hotel in oxford has single or double room? the hotel should has free wifi too} \\ \hline

			restaurant  & 
			\textit{R$_1$ = find all} [\textit{restaurants}] \newline \textit{R$_2$ = find} [\textit{R$_1$}] \textit{with} [\textit{number of reviews}]  \textit{$>$} [\textit{100}] \newline \textit{R$_3$ = count  elements in} [\textit{R$_2$}]&  \textit{how many restaurants have more than 100 reviews}\\ \hline
			restaurant  & 
			\textit{R$_1$ = find all} [\textit{restaurants}] \newline \textit{R$_2$ = find} [\textit{R$_1$}] \textit{with} [\textit{number of reviews}]  \textit{$>$} [\textit{500}] \newline \textit{R$_3$ = find} [\textit{R$_2$}] \textit{with largest number of} [\textit{cuisine}] \newline
			\textit{R$_4$ = find} [\textit{R$_3$}] \textit{which satisfies} [\textit{has outdoor seatings}]        
			&  \textit{for the restaurants with more than 500 reviews, look for those with the largest variety of food, and then those with seats outside}\\ \hline

			disease  &
			\textit{R$_1$ = find all}  [\textit{diseases}] \newline \textit{R$_2$ = find} [\textit{R$_1$}] \textit{with smallest} [\textit{incubation period}] \newline \textit{R$_3$ = find} [\textit{R$_2$}] \textit{where} [\textit{symptom}]\textit{ is not} [\textit{bleeding}] \newline \textit{R$_4$ = count elements in} [\textit{R$_3$}] &  \textit{how many diseases having the smallest incubation period don't result in bleeding}\\ \hline
			
			disease  &
			\textit{R$_1$ = find}  [\textit{symptom}] \textit{of} [\textit{fever}]  \newline \textit{R$_2$ = find all}  [\textit{diseases}] \newline
			\textit{R$_3$ = find} [\textit{R$_2$}] \textit{where} [\textit{symptom}]  \textit{is} [\textit{R$_1$}] \newline
			\textit{R$_4$ = find} [\textit{R$_3$}]	\textit{with largest} [\textit{incubation period}] &  \textit{which disease has the same symptom as fever, and has the longest incubation period}\\ \hline
			
			medication  & \textit{R$_1$ = find all} [\textit{medications}] \newline \textit{R$_2$ = find} [\textit{R$_1$}] \textit{which satisfies} [\textit{for adult only}] \newline \textit{R$_3$ = find} [\textit{R$_2$}] \textit{where} [\textit{target symptom}] \textit{is} [\textit{bleeding}] \newline \textit{R$_4$ = find} [\textit{R$_3$}] \textit{where} [\textit{side effect}] \textit{is} [\textit{headache}]& \textit{what adult medications treat bleeding in exchange for a headache}\\ \hline
			medication  & \textit{R$_1$ = find all} [\textit{medications}] \newline \textit{R$_2$ = find} [\textit{R$_1$}] \textit{where} [\textit{target symptom}] \textit{is} [\textit{headache}] \newline
			\textit{R$_3$ = find} [\textit{R$_2$}] \textit{where}  [\textit{category}] \textit{is} [\textit{physician or pharmacist}] \newline
			\textit{R$_4$ = find} [\textit{R$_3$}] \textit{which satisfies}
			[\textit{requires prescription}] & \textit{find the medicine for headache, with a category of  physician or pharmacist; and the medicine requires prescription}\\ \hline
			
		\end{tabular}
		\caption{Examples of 
			templates (filled values shown within
			brackets), and elicited utterances across six
			domains. $R$ is a shorthand for $Result$. \label{dataexamples}}
	\end{center}
\end{table}

\subsubsection{Comparison to \shortciteA{wang2015building} \label{compar}}
Our data collection method is closely related to
\shortciteA{wang2015building} in that we also ask annotators to
paraphrase artificial expressions into natural sounding ones.  In their
approach, annotators paraphrase a single artificial description.
In comparison, we aim to handle more complex tasks involving
compositional intentions.  For such tasks it is not easy to merge all
meanings into a single formal description.  We therefore represent the
meanings with a sequence of templates, and ask annotators to
summarize the templates into a natural language utterance.

We directly evaluated how annotators perceive our templates compared
to single descriptions, with respect to the compositionality of
meaning representations.  For each domain we randomly sampled~24
meaning representations described by templates (144 tasks in total)
and derived the corresponding artificial description using Wang et
al.'s \citeyear{wang2015building} grammar.  The output from both
approaches was also paired with a task description (manually created
by us) which explained the task (see Table~\ref{comparison} for
examples).  Annotators were asked to rate how well the single artificial
sentence and the templates corresponded to the natural language
description according to two criteria: (a) intelligibility (how easy
is the artificial language to understand?) and (b) accuracy (does it
match the intention of the task?). Participants used a
\mbox{1--5}~rating scale where 1 is~worst and~5 is best. We elicited
5~responses per task.

\begin{table}[t]
	\begin{small}
		\begin{center}
			\begin{tabular}{|@{~}l@{~}|c|@{~}c@{~}|c|c|c|@{~}c@{~}|} \hline
				\multirow{ 2}{*}{Domain}				 & \multicolumn{2}{c@{~}|}{Intelligibility} & \multicolumn{2}{c|}{Accuracy} & \multicolumn{2}{c|}{Combined} \\
				& S & T & S & T  & S & T  \\\hline
				meeting & 3.54 & 3.81 & 3.86 & 4.02 & 3.70 & 3.91 \\
				employee & \underline{3.58} & \underline{4.13} & \underline{3.48} & \underline{4.43} & \underline{3.53} & \underline{4.28} \\
				hotel & 3.81 & 3.96 & \underline{3.79} & \underline{4.29} & \underline{4.12} & \underline{3.80} \\
				restaurant & \underline{3.93} & \underline{4.23} & 3.80 & 3.75 & \underline{3.86} & \underline{4.13} \\
				disease & 3.41 & 3.69 & \underline{3.68} & \underline{4.02} & \underline{3.55} & \underline{3.86} \\
				medication & 3.83 & 3.97 &\underline{3.83} & \underline{4.40} & \underline{3.83} & \underline{4.19} \\
				All & 3.96 & 3.67 & \underline{3.55} & \underline{4.03} & \underline{3.70} & \underline{4.06} \\ \hline
			\end{tabular}
			\caption{Comparison between artificial
                          descriptions (S; Wang et al., 2015) and our
                          template-based approach (T). Mean ratings
                          are shown per domain and overall. Combined
                          is the average of Intelligibility and
                          Accuracy. Means are underlined if their
                          difference is statistically significant at
                          $p<0.05$ using a post-hoc Turk
                          test. \label{compresult}}
		\end{center}
	\end{small}
\end{table}

\begin{table}[t]
	\begin{small}
		\begin{center}
			\begin{tabular}{| c | c | c | c | c | c | c |} \hline
				\multirow{ 2}{*}{Depth} & \multicolumn{2}{c|}{Intelligibility} & \multicolumn{2}{c|}{Accuracy} & \multicolumn{2}{c|}{Combined} \\
				& S & T & S & T  & S & T  \\ \hline
				1 & 4.00 & 4.29 & 4.05 & 4.29 & 4.03 & 4.26 \\
				2 & 4.11 & 4.18 & 4.03 & 4.18  & 4.07 & 4.02  \\
				3 & \underline{3.61} & \underline{4.09} & \underline{3.60} & \underline{4.01} & \underline{3.58} & \underline{4.01} \\
				4 & \underline{3.56} & \underline{4.28} & \underline{3.35} & \underline{4.25} & \underline{3.60} & \underline{4.10} \\
				\hline
			\end{tabular}
			\caption{Comparison between artificial
				sentences (S; Wang et al., 2015) and 
				template-based approach  (T) for varying
				compositionality depths. Mean ratings are
				aggregated across domains. Combined is
				the average of Intelligibility and
				Accuracy. Means are underlined if
				their difference is statistically significant
				at $p<0.01$ using a post-hoc Turk
				test. \label{compresult2}}
		\end{center}
	\end{small}
\end{table}

\begin{table*}[h!]
	\begin{footnotesize}
		\begin{center}
			\begin{tabular}{|@{~}p{5.7cm} |@{~}p{5.7cm} |@{~}p{3.7cm}@{~}|} \hline
				\multicolumn{1}{|c|@{~}}{Task description} &
				\multicolumn{1}{c|@{~}}{Our templates} &
				\multicolumn{1}{c|}{\shortciteA{wang2015building}} \\ \hline
				\amt{We have a database of diseases and would like to
					find  diseases which have fever as their
					symptom. These diseases should be treatable with
					antibiotics. Their incubation period 
					is longer than a day. If you have such
					a disease you should see a doctor.} 
				& \amt{\emph{R}$_{1}$ =   \emph{find the diseases whose
						symptom is fever}} \newline \amt{\emph{R}$_{2}$= \emph{find R$_{1}$
						whose treatment is antibiotics}} \newline \amt{\emph{R}$_{3}$= \emph{find
						R$_{2}$ whose incubation period is longer than a
						day}} \newline \amt{\emph{QR}= \emph{find R$_{3}$ which
						require to see a doctor}} & \amt{diseases whose symptom is
					fever whose treatment is aspirin whose incubation
					period is larger than a day which require to
					see a doctor}\\ \hline
				
				\amt{We have a database of diseases and would like to
					find diseases  which have fever as their symptom;
					amongst them, we would like to find those with heart disease
					as complication. Finally, we want to find all
					diseases that can be treated with antibiotics.} &
				\amt{\emph{R}$_{1}$= \emph{find the diseases whose symptom is fever}}
				\newline \amt{\emph{R}$_{2}$= \emph{find the diseases whose
						complication is heart disease}} \newline \amt{\emph{QR}= \emph{find R}$_{1}$ \emph{and R}$_{2}$ \emph{whose treatment
						is antibiotics}} &\amt{disease whose symptom is fever and
					disease whose complication is heart disease whose
					treatment is antibiotics} \\ \hline
				\amt{We have a database of diseases. We would like to
					first find the incubation period of fever; and
					then find the diseases which have  incubation
					period longer than fever; these diseases can  be also
					treated with antibiotics.} & \amt{\emph{R}$_{1}$=}
				\amt{\emph{find incubation
						period of fever}} \newline \amt{\emph{R}$_{2}$=}
				\amt{\emph{find  diseases
						whose incubation period is larger than R}$_{1}$}
				\newline \amt{\emph{QR= find R}$_{2}$ \emph{whose
						treatment is antibiotics}} & \amt{diseases whose incubation
					period is larger than incubation period of fever
					whose treatment is antibiotics} \\ \hline
			\end{tabular}
			\vspace{-1ex}
			\caption{Templates and artificial sentences
                          \cite{wang2015building} 
				shown to AMT crowdworkers together with task
				description. Examples are taken from the
				disease domain. \emph{R} and \emph{QR} are
				shorthands for \emph{Result} and \emph{Query
					Result}, respectively.
                                      \label{comparison}}
		\end{center}
	\end{footnotesize}
	\vspace{-1ex}
\end{table*}

Table~\ref{compresult} summarizes the mean ratings for each domain and
overall. As can be seen, our approach generally receives higher
ratings for Intelligibility and Accuracy. For both types of ratings,
 templates significantly outperform the individual
descriptions for all domains but meetings. Table~\ref{compresult2} shows
a breakdown of  results according to the depth of
compositionality. Utterances of compositional depth~1 and~2 can be easily described by
one sentence, and our template-based approach has no clear advantage
over \shortciteA{wang2015building}. However, when the compositional depth
increases to~3 and~4, templates are perceived as more intelligible and
accurate across domains; all means differences for depths~3 and~4 are
statistically significant (\mbox{$p<0.01$}).  Further qualitative
analysis suggests that our approach receives higher ratings in cases
where the output of the grammar from \shortciteA{wang2015building}
involves various propositional attachment ambiguities.  The
ambiguities are common when the compositional depth increases (Example
1 in Table~\ref{comparison}), when the query contains conjunction and
disjunction (Example 2 in Table~\ref{comparison}), and when a sub-query
acts as object of comparison in a longer query (Example 3 in
Table~\ref{comparison}).


\subsubsection{Sequential Utterances}
\label{sec:sequ-utter}

We also used AMT to elicit sequential utterances for two domains, namely
restaurant and hotel domains. We simulated a database querying task, where
the user asks a series of questions in order to accomplish an
information need.  Each user session (i.e., querying task) included a
maximum amount of five co-reference, covering
the four co-referential structures described in Section~\ref{multi}.
The Exploitation structure is the building block for all sessions, since
it is the most natural, the user asks follow-up queries to previous
ones. Each session includes an additional structure randomly selected
from Exploration, Merging, and Unrelated, which respectively simulate
comparisons between choices, logical union and intersection, and topic
shift.  In our experiment, these three structures can co-exist with
Exploitation, but not with each other.  More complex co-referential
structures (e.g., two explorations in the same session) are not common
in practice, although they can be constructed from the four basic
structures.

For each domain, we generated meaning representations corresponding to
four types of sessions consisting of Exploitation (only), Exploration,
Merging, and Unrelated.  The number of sessions (described by
templates) for each type was~3,000.  Sessions were sampled randomly
without replacement to show to annotators. Each annotator saw two tasks
per HIT and was paid~0.2\$.  The average amount of time annotators
spent on each domain was three hours.  Finally, we obtained 12,000
examples for each domain.  Table \ref{datastats} shows basic
statistics of our dataset which we call \textsc{Sequential}. These
include input and output vocabulary size, the number of utterances per user
session, the number of tokens per utterance, and the word overlap
between an utterance and its corresponding template.  As can be seen,
\textsc{Sequential} exhibits a significant amount of paraphrasing (more
than half of the tokens in each template are paraphrased).

\begin{table}[t]
	\begin{center}
		\begin{tabular}{ | l | r | r |}
			\hline
			& hotel & restaurant \\ \hline
			Input vocabulary (for natural language) size & 3,813 & 4,628\\
			Output vocabulary (for database) size & 386 & 386\\
			average \#utterances per session & 3.98 & 4.01\\
			average \#tokens per utterance & 9.30 & 9.11\\
			average \#word overlap per utterance-template  & 4.22 & 4.07 \\ \hline
		\end{tabular}
	\end{center}
	\caption{Statistics of the \textsc{Sequential} dataset. \label{datastats}}
\end{table}

The dataset inevitably contains a certain amount of noise, including
spelling mistakes, inaccurate paraphrasing, and wrong
co-reference. However, we did not perform any manual postprocessing as
our goal in this work is to simulate a real-world setting where the
semantic parser has to learn under noise. Nevertheless, we evaluated
100 randomly selected sessions (consisting of~531 queries in total).
Amongst these, 79~sessions are correct, while the remaining~21 contain
wrongly paraphrased utterances.  The paraphrasing accuracy for all the
531 utterances is 94.4\%.  Aside from minor spelling errors, we found
that all mistakes workers made related to co-reference. Most commonly,
they ignored co-reference during paraphrasing. For example, instead of
asking \textsl{of these restaurants, which ones have thai food}, an
annotator may simply write \textsl{which restaurants have thai food}. 

Table~\ref{mdataexamples} and \ref{mdataexamples2} present examples
of the sequential utterances we elicited for the hotel and restaurant
domains together with their co-reference structures and corresponding
templates.

\begin{table}[!ht]
	\begin{center}
		\scriptsize
		\begin{tabular}{|@{~}l | c | p{6cm} | p{6cm} @{~}|} \hline
			\multicolumn{1}{|c|}{Domain}    &   \multicolumn{1}{c|}{Structure}    & \multicolumn{1}{c|}{Templates} &  \multicolumn{1}{c|}{Queries}   \\ \hline
			hotel     & Exploitation &
			\textit{R$_1$ =  find} [\textit{hotels}] \textit{with} [\textit{price rating}] \textit{$\leq$} [\$\$] \newline
			\textit{R$_2$ =  find} [\textit{R$_1$}] \textit{with number of} [\textit{room type}]  \textit{$\geq$} [\textit{4}] \newline
			\textit{R$_3$ =  find} [\textit{R$_2$}] \textit{which satisfies} [\textit{near the sea}] \newline
			\textit{R$_3$ =  find} [\textit{R$_2$}] \textit{which satisfies} [\textit{can be reserved}]
			& 
			\textit{Which hotels have a price rating of 2 dollar signs or less? \newline
				Can you find which of these have more than 4 different types of rooms? \newline
				Among these, which are located near the sea? \newline
				Among these, which takes reservations?} \\ \hline
			
			hotel     & Exploration &
			\textit{R$_1$ =  find} [\textit{hotels}] \textit{with largest number of} [\textit{customer reviews}]  \newline
			\textit{R$_2$ =  find} [\textit{R$_1$}] \textit{with} [\textit{customer rating}]  \textit{$\geq$} [\textit{5 stars}] \newline
			\textit{R$_3$ =  find} [\textit{R$_1$}] \textit{with} [\textit{customer rating}]  \textit{$\geq$} [\textit{3 stars}] \newline
			& 
			\textit{Show me hotels with the largest number of customer reviews \newline
				Which of those hotels have received a customer rating of 5 or more stars? \newline
				Of the hotels with the largest number of reviews, which ones have a customer rating of 3 or more stars?} \\ \hline
			
			hotel     & Merging &
			\textit{R$_1$ =  find} [\textit{hotels}] \textit{with largest number of} [\textit{room type}]  \newline
			\textit{R$_2$ =  find} [\textit{R$_1$}] \textit{with} [\textit{customer rating}]  \textit{$\geq$} [\textit{5 stars}] \newline
			\textit{R$_3$ =  find} [\textit{R$_1$}] \textit{with} [\textit{distance}]  \textit{$\leq$} [\textit{500m}] \newline
			\textit{R$_4$ =  find} [\textit{R$_2$ and R$_3$}] \textit{which satisfies} [\textit{car parks}]  
			& 
			\textit{Which hotel has the most variety of rooms? \newline
				Which of these are rated at 5 stars or more? \newline
				Which of the previous hotels are within 500 meters to me? \newline
				Of all these hotels with 5 stars or near me, which offers car parks?} \\ \hline
			
			hotel     & Unrelated &
			\textit{R$_1$ =  find} [\textit{hotels}] \textit{with largest number of} [\textit{room type}]  \newline
			\textit{R$_2$ =  find} [\textit{R$_1$}] \textit{with smallest} [\textit{price rating}] \newline
			\textit{R$_3$ =  find} [\textit{hotels}] \textit{which satisfies} [\textit{airport shuttle}]  \textit{and} [\textit{private bathroom}] 
			& 
			\textit{What hotels have the largest amount of room types? \newline
				Among those, which have the smallest price rating? \newline
				Which hotels have an airport shuttle and private bathroom?} \\ \hline

		\end{tabular}
		\caption{Examples of co-reference structures,
			templates (filled values shown within
			brackets) and elicited utterances for the hotel
			domain. $R$ is a shorthand notation for $Result$. \label{mdataexamples}}
	\end{center}
\end{table}

\begin{table}[!ht]
	\begin{center}
		\scriptsize
		\begin{tabular}{|@{~}l | c | p{6cm} | p{6 cm} @{~}|} \hline
			\multicolumn{1}{|c|}{Domain}    &   \multicolumn{1}{c|}{Structure}    & \multicolumn{1}{c|}{Templates} &  \multicolumn{1}{c|}{Queries}   \\ \hline				
			restaurant     & Exploitation &
			\textit{R$_1$ =  find} [\textit{restaurants}] \textit{with smallest}  [\textit{price rating}] \newline
			\textit{R$_2$ =  find} [\textit{R$_1$}] \textit{with largest number of} [\textit{food type}]  \newline
			\textit{R$_3$ =  find} [\textit{R$_2$}] \textit{which satisfies} [\textit{good for groups}] \textit{and} [\textit{has waiter service}] \newline
			\textit{R$_4$ =  find} [\textit{R$_3$}] \textit{with} [\textit{distance}] \textit{$\leq$} [\textit{500m}]
			& 
			\textit{Show me the cheapest restaurants. \newline
				Which of these have the largest variety of food? \newline
				Of these, are there any restaurants that are good for groups and offer waiter service?\newline
				Are any of those restaurants within half a kilometer to me?} \\ \hline
			
			restaurant     & Exploration &          
			\textit{R$_1$ =  find} [\textit{restaurants}] \textit{with largest}  [\textit{customer rating}] \newline
			\textit{R$_2$ =  find} [\textit{R$_1$}] \textit{with}  [\textit{distance}] \textit{$\leq$} [\textit{500m}]\newline
			\textit{R$_3$ =  find} [\textit{R$_2$}] \textit{where} [\textit{food type}] \textit{is} [\textit{thai food}] \newline
			\textit{R$_4$ =  find} [\textit{R$_2$}] \textit{where} [\textit{food type}] \textit{is} [\textit{american food}] 
			& 
			\textit{Which restaurants have the largest customer ratings?\newline
				Which of these restaurant is within 500 meters?\newline
				Show me those serves Thai food?\newline
				Of the previous restaurants, find the ones which serves American food instead.} \\ \hline             
			
			restaurant     & Merging &          
			\textit{R$_1$ =  find} [\textit{restaurants}] \textit{where} [\textit{location}] \textit{is} [\textit{downtown}] \newline
			\textit{R$_2$ =  find} [\textit{R$_1$}] \textit{which satisfies}  [\textit{has breakfast}] \newline
			\textit{R$_3$ =  find} [\textit{R$_1$}] \textit{which satisfies}  [\textit{has lunch}] \newline
			\textit{R$_4$ =  find} [\textit{R$_2$} and \textit{R$_3$}] \textit{with largest number of} [\textit{customer reviews}]  
			& 
			\textit{Show me restaurants located downtown? \newline
				Which of these restaurants serve breakfast? \newline
				Which restaurants serve lunch? \newline
				Of all these restaurants with breakfast or lunch, which have the most customer reviews?} \\ \hline            
			
			restaurant     & Unrelated &          
			\textit{R$_1$ =  find} [\textit{restaurants}] \textit{which satisfies} [\textit{can be reserved}] \newline
			\textit{R$_2$ =  find} [\textit{R$_1$}] \textit{which satisfies}  [\textit{take credit card}] \newline
			\textit{R$_3$ =  find} [\textit{R$_2$}] \textit{with number of} [\textit{cuisine}] $>$  [$1$] \newline
			\textit{R$_4$ =  find}[\textit{restaurants}] \textit{with smallest} [\textit{price rating}] \newline
			\textit{R$_5$ =  find} [\textit{R$_4$}] \textit{with largest} [\textit{customer rating}]
			& 
			\textit{which restaurants have reservations option? \newline
				which among these restaurants accept credit cards? \newline
				among these restaurants find those with multiple types of cuisine? \newline
				which restaurants have the smallest price rating? \newline
				which among these restaurants have the largest customer rating?} \\ \hline

		\end{tabular}
		\caption{Examples of co-reference structures,
			templates (filled values shown within
			brackets) and elicited utterances for the restaurant
			domain. $R$ is a shorthand notation for $Result$. \label{mdataexamples2}}
	\end{center}
\end{table}

\subsubsection{Can Annotators Create Their Own Tasks?}
\label{sec:dynam-mode-eval}

In the data collection methods described so far, annotators are shown
templates and asked to write down natural language expressions
summarizing them. The fact that templates are generated automatically
allows us to explore a potentially very large space of underlying
meaning representation exhibiting compositionality and wide coverage.
However, in practice, companies sometimes hire annotators who are in
charge of creating domain-specific tasks on their own.  These annotators
are often trained with domain knowledge but lack in programming knowledge.  
We would like to further explore if annotators are able to create their own querying tasks and annotations, with the aid of templates.

Since our data elicitation method maps each domain-general rule into
a template, which is both human-readable and machine interpretable, we
can give annotators the freedom to manipulate templates for generating
their own tasks.  In order to specify a task, annotators need to
selectively use a sequence of templates and instantiate them with
domain-specific information.  While they do so with an annotation tool
at the front end, meaning representations are created automatically at
the back end.

We conducted an experiment to explore this idea.  Because it is not
realistic to recruit a large number of domain experts to participate
in our evaluation, we ran this experiment on AMT with annotators who
were paid more to compensate for the increased workload.  The
experiment was conducted on the same six domains (meeting, employees,
hotel, restaurant, disease and medication) used to elicit the
\textsc{SingleTurn} dataset.  For each domain, annotators were given
unfilled templates and a table describing naturalized predicates and entities of
that domain. They were also given instructions and examples explaining
how to use the templates and domain information.  Annotators were then
asked to chose the templates, fill them, and write down an utterance
summarizing the task.  We recruited 50 workers for each domain (300 in
total) and each was asked to complete two tasks in one HIT worth~1\$.

\begin{table}[t]
	\begin{small}
		\begin{center}
			\begin{tabular}{|l|c|c|c|c|} \hline
				Domain    & Tk & WO  & Tp & Acc   \\ \hline
				meeting   & 16.71 & 2.83 & 3.46 & 0.925 \\
				employees  & 18.64 & 3.71 & 3.33 & 0.938 \\
				hotel     & 16.97 & 4.08 & 3.37 & 0.969 \\
				restaurant  & 17.33 & 3.72 & 3.36 & 0.943 \\
				disease  & 18.95 & 5.12 & 3.12 & 0.925 \\
				medication   & 17.25 & 3.26 & 3.15 & 0.989\\ \hline
			\end{tabular}
			\caption{Average number of tokens (Tk), token
                          overlap between queries and templates (WO),
                          average number of templates (Tp) and
                          proportion of parsable templates (Acc) per
                          domain when annotators are given flexibility
                          to generate querying tasks.  \label{dmevaluation}}
		\end{center}
	\end{small}
	\vspace{-1ex}
\end{table}

Table \ref{dmevaluation} shows the statistics of the data we obtained.
Workers tend to use more than three templates (on average) per task,
and the degree of paraphrasing is increased compared to the original
data collection method where tasks are generated automatically.
Moreover, across domains, 90\%~of the meaning representations created
by AMT annotators are executable.  Although we envisage domain experts as
the main users of such an annotation tool, the result indicates that (with
proper instructions and examples as well as adequate pay) regular AMT
annotators can also generate meaningful meaning representations to some
extent.  Inspection of the output failures revealed two common
reasons. Firstly, annotators filled in templates with wrong types. For
example, one worker filled the \texttt{LookupKey} template with an
entity (e.g.,~\textit{find all} [\textit{annual review}]) instead of a
database key; and another worker used \texttt{Count} as the first
template followed by \texttt{Filter}, however type constraints require
\texttt{Filter} to take a set of entities as argument instead of a
number (as returned by \texttt{Count}). Secondly, annotators ignored
information in the table and created tasks using non-existing domain
information.  Since meaning representations can be type-checked for
validity automatically, providing feedback on the fly may yield a
higher percentage of executable meaning representations.

\subsection{Semantic Parsing}
In this section, we evaluate our neural semantic parser
(Section~\ref{nsp}) on the two datasets we collected
(\textsc{SingleTurn} and \textsc{Sequential}).

\subsubsection{Semantic Parsing for Single-turn Utterances}
\label{sec:semant-pars-results}
In the single-turn experiments, our parser parses an utterance into a meaning
representation which does not contain co-referential
variables\footnote{During the construction of the meaning
representation, co-referential variables are always replaced with their
values---which are antecedent meaning representations.}.  

Our parser was trained on the \textsc{SingleTurn} dataset presented in
Section~\ref{sec:data-collection}. Training/validation/testing splits are 0.7, 0.1 and 0.2. All LSTMs had one layer with
150~dimensions. The word embedding size and rule embedding size were
set to~50.  A dropout of~0.5 was used on the input features of the
softmax classifiers in Equations~(13)--(15) (and (16)--(19)).
Momentum SGD was used to update the parameters of the model.  We
compared our model with two baselines, a sequence-to-sequence (S2S)
parser \cite{dong2016language,jia2016data} which converts an utterance
to a meaning representation in string format, and a sequence-to-tree
(S2T) parser which converts an utterance directly to an output tree
\cite{cheng2017nsp}.  As an example, S2S generates the meaning
representation \texttt{(filter (Result$_1$) (rel.distance) < (num
  500))}, S2S from left to right including auxiliary brackets
\texttt{(} and \texttt{)}.  S2T generates the same meaning
representation as a tree using cues from brackets:
\texttt{Result$_1$}, \texttt{rel.distance}, \texttt{<}, and
\texttt{(num 500)} are all attached as children nodes to non-terminal
\texttt{filter}.  Note that S2T is a reasonable model for recursive
meaning representations, whose structure reveals how meaning
representations are compositionally obtained.  However, the generation
process of S2T is not interpretable for non-recursive meaning
representations.  Different from S2T, our model (denoted by S2D)
generates the representation for the same example with a non-terminal
rule \texttt{filter(\texttt{<})}, and three terminal rules
\texttt{Coref}, \texttt{BinaryPredicate} and \texttt{Entity}.  These
rules are then composed to derive the final logical form.

\begin {table*}[t]
\begin{center}
	\small
	\begin{tabular}{|l | ll | ll |  ll | ll |  ll | ll |}
		\hline
		\multirow{ 2}{*}{Model}   &  \multicolumn{2}{c|}{meeting}	&
		\multicolumn{2}{c|}{employees}	&
		\multicolumn{2}{c|}{hotel} &
		\multicolumn{2}{c|}{restaurant} &
		\multicolumn{2}{c|}{disease}	&
		\multicolumn{2}{c|}{medication   }\\ 
		&          ExM & SeM & ExM & SeM & ExM & SeM & ExM & SeM & ExM & SeM & ExM & SeM \\\hline
		S2S  & 37.2 & 43.2 & 14.3 & 17.5 & 24.5 & 31.2 & 21.3 &
		29.0 & 16.7 & 23.2 & 15.5 & 16.7\\
		S2T  & 41.2 &  46.8 &21.4 & 28.2 & 31.5 &  43.5 & 25.4 &
		35.9 & 22.4 & 34.4 & 28.2 & 33.9\\
		S2D & 45.6 & 54.0 & 27.8 & 35.5 & 39.2 & 52.6 & 47.2 &
		49.5 & 26.9 & 44.6 & 35.8 & 46.2\\ \hline
	\end{tabular}
\end{center}
\caption{Performance on various domains (test set) using exact match (ExM) and
	semantic match (SeM). \label{result_match}}
\end{table*}

Table~\ref{result_match} shows results on the test set of each domain
using exact match as the evaluation metric (ExM).  Our
sequence-to-derivation model (S2D) yields substantial gains over
the baselines (S2S and S2T) across domains.  However, a limitation of
exact match is that different meaning representations may be
equivalent due to the commutativity and associativity of rule
applications. They can eventually execute to the same denotation.  For
example, two subsequent \texttt{Filter} rules in a meaning
representation are interchangeable.  For this reason, we additionally
compute the number of meaning representations that match the gold
standard at the semantic or denotation level.   Again, we find that the S2D
outperforms related baselines by a wide margin.  This result is not
surprising, since S2D explicitly models the compositionality of
meaning representations.

We conducted further experiments by training a single model on data
from all domains, and testing it on the test set of each individual
domain.  As the results in Table~\ref{result_multi} reveal, we obtain
gains for most domains on both metrics of exact and semantic match.
Our results agree with previous work \cite{herzig2017neural} which
improves semantic parsing accuracy by training a single sequence-to-sequence model over multiple domains.  Since domain-general
aspects are shared in our model, cross-domain training offers our parser
more supervision cues.

\begin{table}[t]
		\begin{center}
			\begin{tabular}{|l|c|c|} \hline
				\multicolumn{1}{|c|}{Model}    & ExM & SeM   \\ \hline
				meeting   & (45.6) 48.8& (54.0) 56.8\\
				employees  & (27.8) 31.7 &  (35.5) 41.4 \\
				hotel     & (39.2) 41.8& (52.6) 56.1\\
				restaurant  & (47.2) 33.1&  (49.5) 48.8\\
				disease  & (26.9) 30.7&  (44.6) 48.3\\
				medication   & (35.8) 37.4& (46.2) 51.8\\ \hline
			\end{tabular}
			\caption{Sequence-to-derivation tree model
                          (S2D) trained on all six \textsc{SingleTurn}
                          domains and evaluated on the test set of
                          each domain. Results of S2D when trained and
                          tested on a single domain are shown within
                          brackets. \label{result_multi}}
		\end{center}
\end{table}


\subsubsection{Semantic Parsing for Sequential Utterances}
Next, we evaluate our neural semantic parser on sequential utterances.
Specifically, each utterance in a user session is paired with a meaning
representation which can contain co-referential variables.  The goal
of the parser is to parse every utterance in the session.  

Our semantic parser was trained and test on the \textsc{Sequential}
dataset introduced in Section~\ref{multi}.  We adopt the same experimental setup as in
our single-turn experiments. Training/validation/testing splits are 0.7, 0.1 and 0.2. All LSTMs have one layer with
150~dimensions, and all word and rule embeddings have size~50.  Recall
that we use a self-attention network to compute the value of
co-referential variables.  This network relies on a bidirectional-LSTM
to compute utterance vectors.  In our experiments, the weights of this
bidirectional-LSTM are shared with the encoder of the semantic parser.
The encoder bidirectional-LSTM computes a list of token vectors,
instead of a single vector per utterance.  We compare this parser
(S2D) against a context dependent parser (S2D$_{\mathcal{C}}$) and two
baselines, a vanilla sequence-to-sequence models (S2S) and
sequence-to-tree (S2T) model.

For evaluation, we primarily consider the parsing accuracy of each
individual utterance (ExM).  This accuracy is first averaged within
each user session and then averaged over all sessions.  Similar to the single-turn
evaluation, we additionally measure the match at the
semantic or denotation level (SeM).  We also break down our results
based on data sessions involving different co-referential structures:
with Exploitation only, and Exploitation combined with
Exploration/Merging/Unrelated.
 
\begin{table}[t]
		\begin{center}
			\begin{tabular}{ |ll ll | ll| ll | ll | ll |  ll |}
				\hline
&\multirow{ 2}{*}{Model}	& \multicolumn{2}{c}{Exploitation} & \multicolumn{2}{c}{Exploration} & \multicolumn{2}{c}{Merging} & \multicolumn{2}{c}{Unrelated}  & \multicolumn{2}{c|}{All} \\
 &&ExM & SeM  &   ExM & SeM    & ExM & SeM   & ExM & SeM  & ExM & SeM  \\\hline
&				 \textsc{S2S} & 46.9 & 23.8& 46.2&
                                 23.2 &38.3 & 3.10  & 38.3 & 3.80 &
                                 42.0 & 44.5 \\
&				 \textsc{S2T} & 69.2 & 40.7 &  64.1
                                 &45.5 & 50.9 &4.70 & 50.4 &6.30 &
                                 57.9 & 61.2 \\
&				 \textsc{S2D} & 72.5 & 42.5 & 67.7 &
                                47.1 &  53.1 & 5.70 &  54.3  & 7.00&
                                60.2 & 25.1 \\
\raisebox{.02cm}[0pt]{\begin{sideways}restaurant\end{sideways}}&				 \textsc{S2D}$_{\mathcal{C}}$ & 
                                  73.6 & 42.8 & 68.1 & 47.1 & 54.6 &
                                  5.80 & 54.9 & 7.1 & 62.4 & 25.1\\
				\hline
&\textsc{S2S} &  48.3 & 24.6 & 48.2 & 19.5& 43.6 & 6.3& 41.5 & 4.10 &  44.5& 13.6\\
&\textsc{S2T}& 66.9 & 42.1 & 72.3& 54.7& 62.4 & 9.6&53.8&5.40 & 61.2& 25.2 \\
&\textsc{S2D} & 73.8 & 43.7 &  77.8& 56.8& 65.4 & 11.8& 57.2& 6.70 &  66.9&29.7\\
\raisebox{.4cm}[0pt]{\begin{sideways}hotel\end{sideways}}&\textsc{S2D}$_{\mathcal{C}}$ & 73.7 & 43.6 & 78.8 & 56.8& 65.0&
11.8&59.5& 7.10& 67.0&29.8\\ \hline

			\end{tabular}
		\end{center}
	\caption{Sequential semantic parsing results on the restaurant and hotel domains. \label{mresults}}
\end{table}

Table~\ref{mresults} shows our semantic parsing results, which clearly
reveal the advantage of S2D over S2S and S2T.  We see that results for
semantic match are rather low, indicating that it is challenging to
parse all utterances in a session correctly.  Comparing among various
sessions, those with Merging and Unrelated result in lowest
accuracy. To better understand how the model does on Exploration,
Merging and Unrelated, we evaluate parsing performance (precision,
recall, and F1) for specific utterances involving these structures.

For the Exploration structure, precision, recall and F1 are 84.4\%,
73.6\%, 78.6\% respectively for the restaurant domain (87.7\%, 71.9\%,
79.0\% for the hotel domain).  For the Merging structure (which involves
union or intersection), precision, recall, and F1 are 88.6\%, 21.4\%,
and 34.5\%, respectively for the restaurant domain (94.8\%, 43.2\%,
and 59.4\% for the hotel domain).  The low recall indicates that most
of the union or intersection symbols in the meaning representations
are not discovered correctly.  Finally, in the Unrelated structure, we
have two independent querying tasks in one session.  We predict how
the model performs on predicting the first utterance of the second
querying task, which indicates the model's ability to detect topic
shift.  Precision, recall, and F1 are 81.9\%, 63.8\%, and 71.7\%,
respectively for the restaurant domain (82.8\%, 68.1\% and 74.7\% for
the hotel domain).  Overall, we see that the parser does fairly well
on Exploitation, Exploration, and Unrelated, while it fails to
recognize most Merging structures.


Table~\ref{mresults} also shows the results of the context-dependent
semantic parser (\textsc{S2D}$_{\mathcal{C}}$). Across different co-reference structures, modeling
context explicitly during decoding leads to some performance improvements;
however, the gains are relatively small. We think this is due to the
fact that a single utterance often provides enough cues to infer an
explicit or implicit co-referential variable.  Explicit co-reference
is evidenced by pronouns (e.g., ``\textsl{find restaurants in the
  oxford street?}''  followed by ``\textsl{those with car parks?}''),
while in utterances with implicit co-reference the querying object is
often missing (e.g., ``\textsl{find restaurants in the oxford
  street?}''  followed by ``\textsl{with car parks?}'').  It is not
difficult for a non-context dependent neural network to learn these
surface cues from training data.  An exception is when an utterance
contains nominal anaphora (e.g., ``\textsl{find restaurants in the
  oxford street?}''  followed by ``\textsl{the oxford restaurants with
  car parks?}'').  In this case, modeling context helps the parser
identify a co-referential relation in the consequent utterance, where
``\textsl{oxford}'' refers to ``\textsl{oxford street}'' and
``\textsl{the oxford restaurants}'' refers to the denotation of the
antecedent utterance.

\section{Conclusions \label{conclusion}}

In this work we examined a framework of building a neural semantic parser from a domain ontology.
We focus on compositional tasks that represent  complex human intentions.
First, our data
elicitation method starts from the formal language space,
which computers are able to explore to generate meaning
representations.  These meaning representations are 
mapped to templates which are finally converted by humans into single or sequential
utterances.  Next, our neural semantic parser leverages the annotations we obtain to
generate derivation trees of
meaning representations, by explicitly modeling the composition of
rules.
Experimental results show that our framework provides an end-to-end solution for building a
neural semantic parser from a domain ontology. This is useful when a new domain is introduced or
 a domain ontology is updated.

 Future work will focus on devising
 smarter data generation methods with less human involvement.  For
 example, natural language generation from formal descriptions may be
 automated with a statistical generative model.   Besides, we 
would like to apply similar data elicitation-modeling  ideas to build conversational assistant \cite{liang2018data}, 
which involves dialog flow management and decision making for optimum response strategies.


\bibliographystyle{theapa}
\bibliography{sample}

\end{document}